\documentclass[]{fairmeta}
\usepackage[T1]{fontenc}

\usepackage[utf8]{inputenc}

\usepackage{microtype}

\usepackage{graphicx}
\usepackage{booktabs}
\usepackage{subcaption}
\usepackage{multirow}
\usepackage{amsmath}
\usepackage{amssymb}
\usepackage{fvextra}
\usepackage{xspace}
\usepackage{newfloat}
\usepackage{colortbl}
\usepackage{tcolorbox}
\usepackage{wrapfig}
\usepackage{needspace}
\tcbuselibrary{breakable,skins}

\DeclareFloatingEnvironment[name=Algorithm]{algorithm}
\newcommand{\model}{SeVeR\xspace}
\newcommand{\dataset}{{BreMRIs-VQA}\xspace}

\title{\model: Selective Visual Exposure and Retrieval for 3D Medical Image Question Answering}

\author[1,2,5,6,7,*]{Yaojun Hu}
\author[1,3,*,\ddagger]{Danyang Tu}
\author[1,*]{Yang Liu}
\author[1,3,\dagger]{Jiajin Zhang}
\author[1,3]{Wei Fang}
\author[2]{Zhiqiang Liu}
\author[2,5,6,7]{Chunlai Dong}
\author[1]{Yingda Xia}
\author[4,5,6\dagger]{Haochao Ying}
\author[2,5,6,7]{Jian Wu}
\author[1]{Ling Zhang}

\affiliation[1]{DAMO Academy, Alibaba Group\\}
\affiliation[2]{College of Computer Science and Technology, Zhejiang University\\}
\affiliation[3]{Hupan Lab\\}
\affiliation[4]{School of Public Health, Zhejiang University\\}
\affiliation[5]{State Key Laboratory of Transvascular Implantation Devices and TIDRI, Zhejiang University\\}
\affiliation[6]{The Second Affiliated Hospital and Liangzhu Laboratory, Zhejiang University School of Medicine\\}
\affiliation[7]{Zhejiang Key Laboratory of Medical Imaging Artificial Intelligence}

\contribution[*]{Equal contribution}
\contribution[\ddagger]{Project leader}
\contribution[\dagger]{Corresponding author}

\abstract{
Volumetric medical VQA requires reasoning over long and redundant 3D visual token sequences, especially in multi-sequence MRI where complementary modalities provide diverse diagnostic cues but expose the decoder to many repeated anatomical regions. To investigate reasoning under multi-sequence visual redundancy, we first introduce \textbf{BreMRIs-VQA}, a clinically curated breast MRI benchmark with 1.19M QA pairs from 71.0K sequences and 12.9K patients, covering both free-text and multiple-choice questions. We further propose \textbf{\model}, a selective visual exposure framework that compresses dense volumes into modality-wise prototypes and retrieves complementary multi-level evidence with change-aware gated attention during decoding, trained with a marginal-utility self-consistency objective that suppresses unhelpful retrieval. Experiments on BreMRIs-VQA and public benchmarks show that \model improves both discriminative and generative performance while exposing substantially fewer visual tokens. 
} 
\begin{document}
\thispagestyle{firstheader}
\maketitle
\pagestyle{empty}

\section{Introduction}

Medical Visual Question Answering (VQA) is evolving from recognition in single images to answering clinically grounded queries over complex imaging studies~\cite{li2023llavamed, lin2025healthgpt, xu2025lingshu}. This challenge is particularly pronounced in volumetric MRI, where diagnosis depends on integrating complementary evidence across multiple sequences. For example, a lesion with suspicious enhancement on dynamic contrast-enhanced MRI may be interpreted differently after considering its T2-weighted signal and diffusion restriction pattern~\cite{mri_basics}. Therefore, an effective medical VQA model needs to integrate complementary evidence distributed across multi-sequence 3D images. However, existing methods~\cite{hamamci2024ct2rep, wu2025towards, bai2024m3d} still predominantly rely on single-modal inputs, overlooking the complementary interactions inherent in multi-sequence MRI. 

\begin{figure*}
    \centering
    \includegraphics[width=0.98\linewidth]{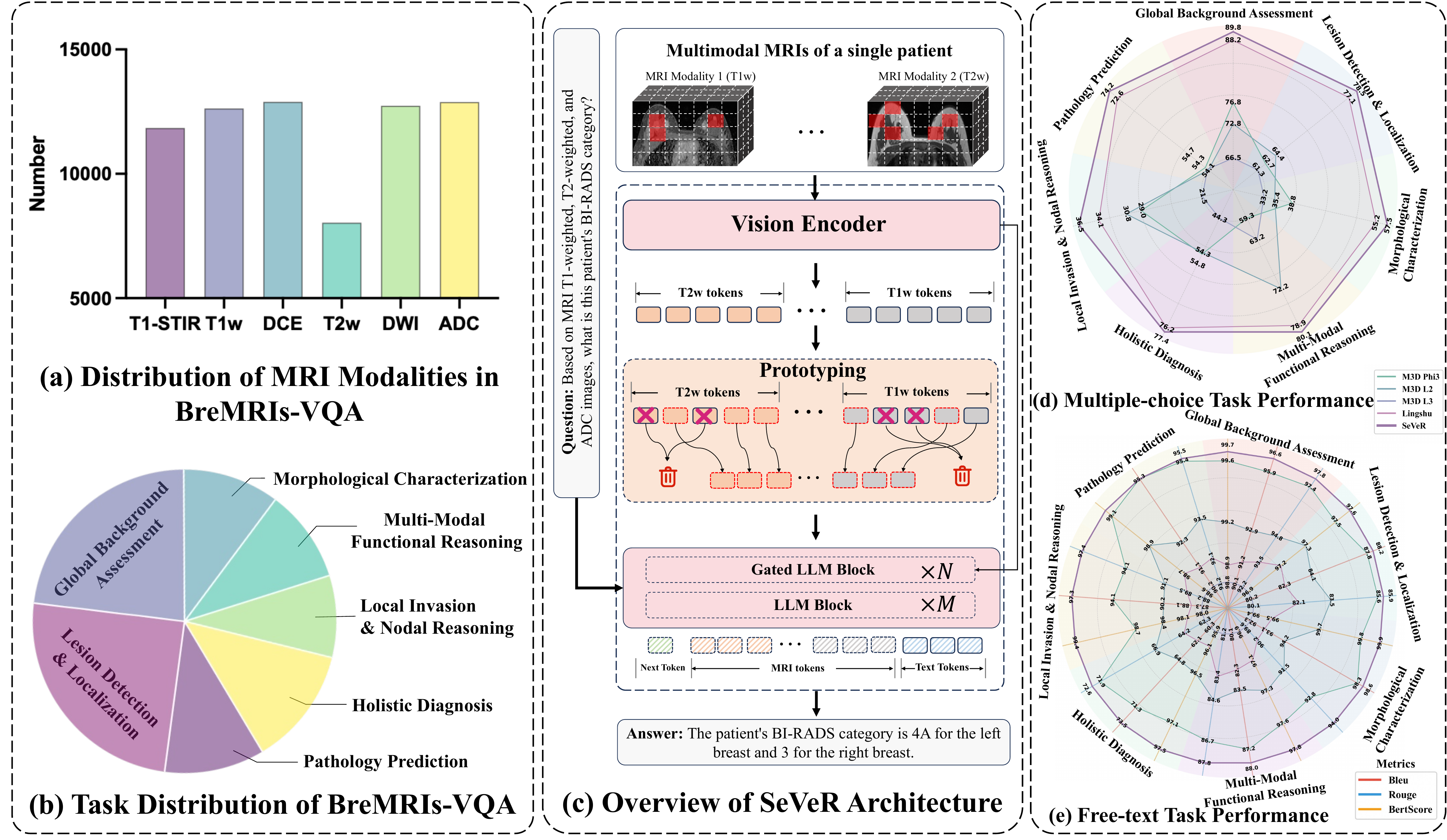}
    \caption{Overview of the BreMRIs-VQA benchmark and the proposed \model. \textbf{(a--b)} BreMRIs-VQA contains 6 MRI modalities and 7 categories of clinical tasks derived from 1.19M QA pairs. \textbf{(c)} Architecture of \model, which integrates Greedy Prototype Selection, Change-aware Gated Attention, and Self-Consistency Regularization. \textbf{(d--e)} \model's superior performance in multiple-choice and free-text generation tasks.}
    \label{fig1}
\end{figure*}

In this paper, we try to advance existing single-modality-based medical VQA methods to the multi-modal scenario, allowing a better alignment with clinical practice. To this end, two challenges have emerged: \textit{1) Scarcity of multi-modality-based medical VQA datasets}: due to the inherent inaccessibility of medical data and the requirement for clinical expertise in designing multi-modality based question-answer (QA) pairs, existing medical VQA datasets~\cite{gai20253drad, chen2025deeptumor}, typically contain only single-modal image and thus feature relatively limited types of QA pairs; \textit{2) Redundant visual exposure}: in multi-modal scenario, images of different modalities contain complementary diagnostic signals while simultaneously sharing extensive common regions, leading to the challenge of signal dilution ~\cite{Chen2024fastv,bolya2023token, yang2025visionzip}, \textit{i.e.}, salient findings in one single modality may be overwhelmed by repetitive background patterns of all other modalities. 

To tackle the data challenge, we first built \textbf{BreMRIs-VQA}, a large-scale, clinically curated multi-modal medical VQA benchmark dataset. We collected \textbf{71.0K} breast MRI sequences from a cohort of \textbf{12.9K} patients, where each contains at least 3 different MRI modalities, paired with a radiology report and a pathology report verified by two radiologists. Then, we design a QA generation pipeline consisting of multi-stage extraction, task construction, paraphrasing, and validation,  which enables BreMRIs-VQA to contain \textbf{1.19M} VQA pairs, including \textbf{671.6K} free-text pairs and \textbf{515.1K} multiple-choice pairs. 
The benchmark covers seven workflow-grounded task groups, enabling evaluation of both generative reasoning and discriminative clinical decision making. Fig.~\ref{fig1} \textbf{(a)} illustrates the number of 6 MRI modalities (T1-STIR, T1w, DCE, T2w, DWI and ADC) included the dataset, while Fig.~\ref{fig1} \textbf{(b)} depicts totally 7 VQA task categories ranging from fine-grained lesion characterization to holistic diagnosis. 

We further propose \textbf{\model}, a selective visual exposure framework for multi-sequence medical VQA, as shown in Fig.~\ref{fig1} \textbf{(c)}. \model first applies \emph{Greedy Prototype Selection (GPS)} to compress dense volumetric tokens into compact modality-wise prototypes that preserve global coverage while suppressing repeated visual content. Since fixed prototypes may miss question-specific details, \model maintains a multi-level feature bank and uses \emph{Change-aware Gated Attention (CaGA)} to retrieve complementary evidence during decoding. This design separates two roles: GPS provides a compact visual memory before decoding, while CaGA performs question-dependent evidence retrieval through the language-conditioned decoder states. To avoid degenerate always-on retrieval, we introduce \emph{Self-Consistency Regularization with Marginal Utility (SCR-MU)}, which penalizes retrieval when it fails to improve task loss over a retrieval-disabled baseline. Figs.~\ref{fig1} \textbf{(d)} and \textbf{(e)} provide an overview of the resulting gains across multiple-choice and free-text clinical tasks.

Our contributions are threefold. First, we introduce \textbf{\dataset}, a large-scale multi-sequence breast MRI VQA benchmark with workflow-grounded free-text and multiple-choice questions from expert verified clinical reports. 
Second, we propose \textbf{\model}, which initially selects compact representative visual prototypes and dynamically retrieves fine-grained evidence via change-aware attention gates with a marginal-utility self-consistency objective that prevents collapsing to trivial behaviors. 
Finally, extensive experiments on \dataset and public 3D benchmarks validate the effectiveness of the proposed framework across discriminative and generative tasks.

 \begin{figure*}[t]
    \centering
    \includegraphics[width=0.88\linewidth]{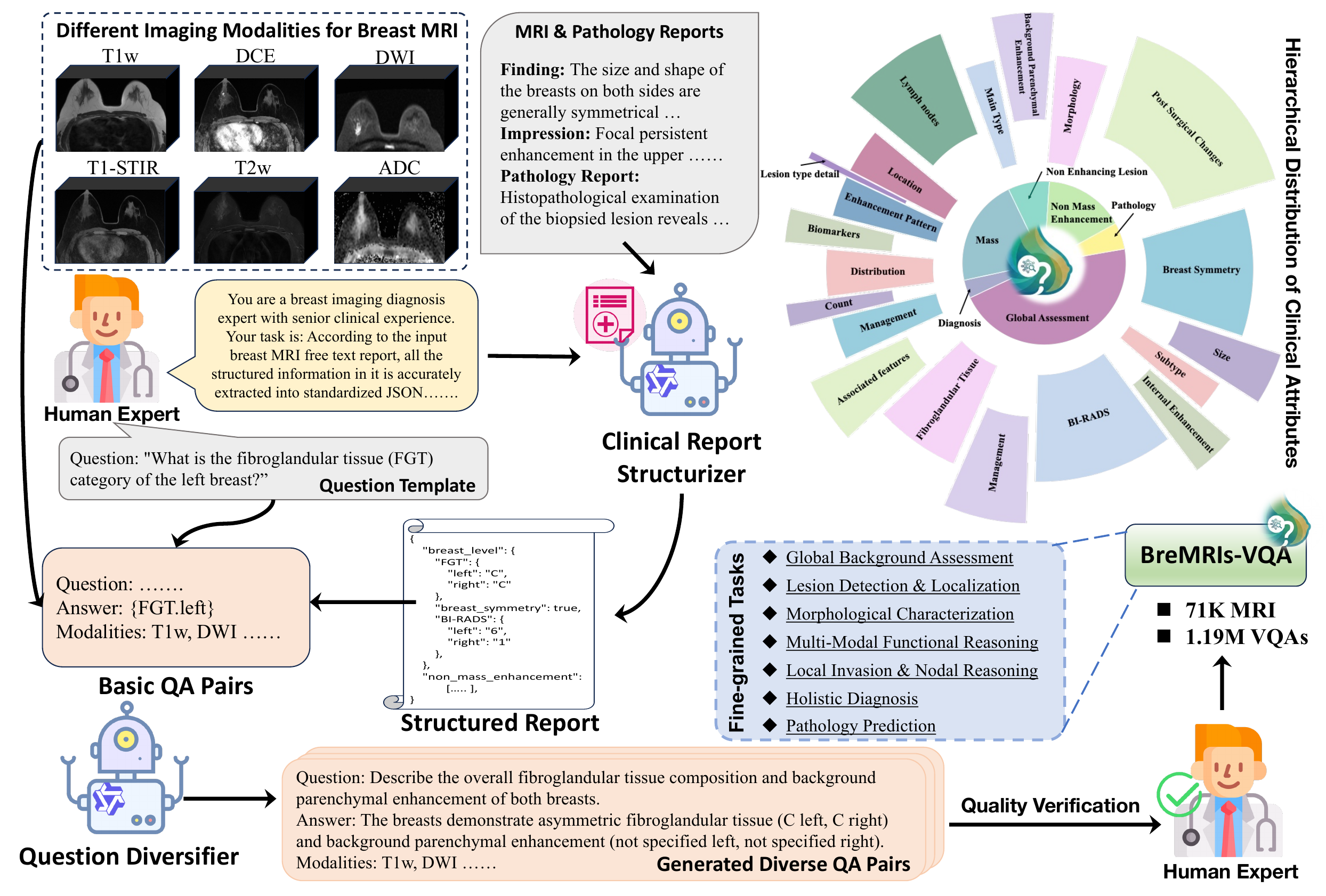}
    \caption{Dataset construction pipeline. The top-right pie chart depicts hierarchical clinical attribute distribution.} 
    \label{fig:dataset}
\end{figure*}
\section{\dataset}

Existing medical VQA datasets are dominated by single-modality settings, limiting evaluation of multi-sequence clinical reasoning. To address this gap, \dataset is built from 71,041 breast MRI sequences of 12,891 patients, with at least three sequences per patient and heterogeneous availability across six MRI modalities: T1w, T1-STIR, T2w, DWI, ADC, and DCE. Fig.~\ref{fig1}(a) summarizes the modality distribution. These imaging studies are paired with \textit{clinical-expert-verified} radiology and pathology reports, which provide the evidence sources for QA construction. The released benchmark contains \textbf{1,186,726} volumetric QA pairs, including \textbf{671,596} free-text and \textbf{515,130} multiple-choice pairs, spanning seven workflow-grounded tasks. Evaluation uses the case-level held-out split detailed in Appendix~\ref{app:bremris_dataset_statistics}. Ethics, privacy, and governance are reported in Appendix~\ref{app:ethics}.

\vspace{-0.5em}
\subsection{Dataset Construction}
We build \dataset with a three-stage LLM-assisted but evidence-constrained pipeline (Fig.~\ref{fig:dataset}), where the LLM never invents clinical facts from images but only structures expert-verified reports and diversifies report-grounded questions.

\noindent\textbf{Stage I: Structured extraction.}
We use Qwen3-max~\cite{yang2025qwen3} with clinician-designed prompts to normalize radiology and pathology reports into a JSON schema with a closed clinical vocabulary (e.g., FGT $\in$ \{A,B,C,D\}), covering breast-level findings, lesion attributes, multi-sequence imaging signals, BI-RADS-related factors, and pathology. Fields not mentioned in the report are set to \texttt{null}, and schema-invalid or out-of-vocabulary outputs are discarded before question generation. Full schemas and prompts are in Appendix~\ref{app:dataset_construction_detail}. 

\noindent\textbf{Stage II: Question generation.}
We instantiate questions from deterministic \textit{clinician-designed templates}, each tied to a source key, answer space, and task label. The templates cover seven workflow-grounded tasks: global background assessment, lesion detection/localization, morphological characterization, multi-modal functional reasoning, local invasion/nodal reasoning, holistic diagnosis, and pathology prediction. This organization supports capability-specific evaluation while preserving clinical interpretability. Full templates and key--task mappings are in Appendix~\ref{app:stage2_task}.

\noindent\textbf{Stage III: Paraphrasing and filtering. }
To improve linguistic robustness, we use Qwen3-max to diversify template-generated questions, after the source key, answer label, and answer space have already been fixed by \textit{deterministic templates}. This step does not create new clinical facts or change supervision labels. Each candidate is re-bound to its template and rejected if it queries a different attribute, leaks the gold answer or its synonyms (e.g., ``BI-RADS 4A'' $\leftrightarrow$ ``category 4A'') into the question, or maps the answer outside the predefined answer space. Surviving paraphrases inherit the original label verbatim.
\begin{figure*}[t]
    \centering
    \includegraphics[width=0.99\linewidth]{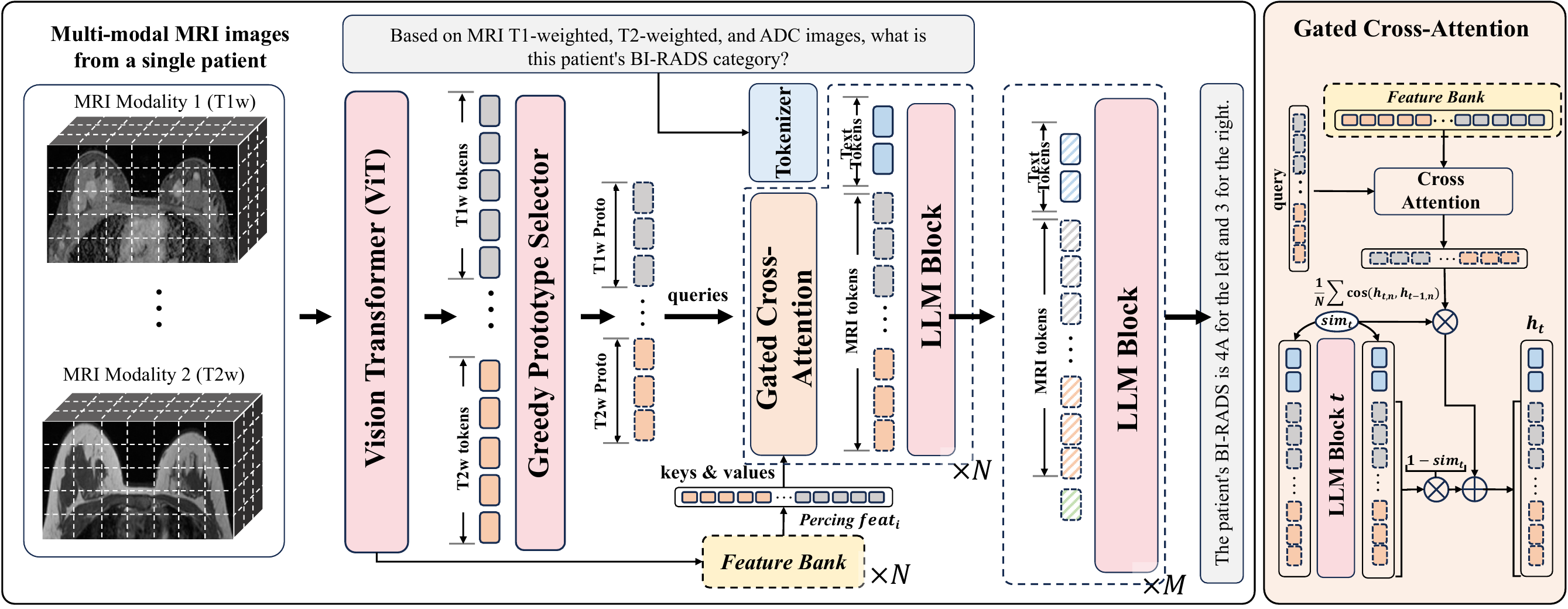}
    \caption{Overview of \model. Multi-modal 3D volumes are first encoded into dense tokens via a shared 3D ViT. The Greedy Prototype Selector (GPS) distills redundant tokens into compact prototypes using global affinity to preserve broad visual coverage. Change-aware Gated Attention (CaGA) dynamically injects multi-level features from a bank during decoding, using language-conditioned decoder states to retrieve complementary evidence.}
    \label{fig:method_overview}
\end{figure*}
\subsection{Quality Control and Verification}
To ensure factual grounding, each QA pair is scored by Qwen3-Coder-Plus on factual consistency, clinical validity, and overall report support, and candidates with weak support are discarded or escalated for manual review. In parallel, radiologists audit 3{,}500 stratified QA pairs (500 per task) for answer--evidence pass rate. The retained set achieves an average overall score of 2.85/3 and a 97.5\% expert pass rate across tasks (Table~\ref{tab:quality_control}), and a closed-form answer-leakage filter further removes paraphrases whose question reveals the gold answer or its clinical synonyms. Full protocols are in Appendix~\ref{app:dataset_qc_detail}.

\section{Method}

For each patient, we consider a set of multi-modal 3D medical image volumes, denoted as $\{\mathcal{V}^{m}\}_{m=1}^{M}$, where each $\mathcal{V}^{m} \in \mathbb{R}^{Z \times H \times W}$ represents a distinct imaging modality. 
We first encode the 3D volume of each modality independently using a shared 3D Vision Transformer. 
After flattening along the spatial dimensions, we get a dense token set $\{X^{m}  \in \mathbb{R}^{L_m \times d}\}_{m=1}^M$, where $L_m$ represents the number of patches in the $m$-th modality.

\subsection{Greedy Prototype Selector}
\label{sec:grps}

As illustrated in Fig.~\ref{fig:method_overview}, processing all visual tokens from multi-sequence volumes leads to redundant visual exposure. We therefore use a Greedy Prototype Selector (GPS) to initialize a compact \emph{question-agnostic visual memory} for each modality, preserving broad coverage of the modality-specific feature manifold under a small token budget. Unlike MMTok~\cite{dong2026mmtok}, where coverage maximization serves as a standalone token selection strategy, GPS in \model only prepares the visual candidate space.
Specifically, we first $\ell_2$-normalize all token features in the $m$-th modality $X^{m}$ to obtain $\{\tilde{\mathbf{x}}^{m}_i\}_{i=1}^{L_m}$ and compute the pairwise affinity matrix $A$, where $A_{ij}^m=(\tilde{\mathbf{x}}_i^m)^\top\tilde{\mathbf{x}}_j^m$. 

A temperature-scaled row-wise softmax is then applied to yield the normalized affinity matrix:
\begin{equation}
\hat{A}_{ij}^m=\frac{\exp(A_{ij}^m/\tau_v)}{\sum_{j'=1}^{L_m}\exp(A_{ij'}^m/\tau_v)} ,
\end{equation}
where $\tau_v$ controls the sharpness of the affinity distribution. To prevent self-dominance during selection, we set the diagonal elements $A_{ii}^m=-\infty$.
With this affinity map, we can identify a compact subset of tokens that effectively covers the modality-specific feature manifold. First, we define the coverage score of a candidate subset $S^m \subset X^{m}$ as
\begin{equation}
f(S^m, X^{m})=\frac{1}{L_m}\sum_{i=1}^{L_m}\max_{j\in S} \hat{A}_{ij}^m.
\label{eq:facility}
\end{equation}
Then, we get $k$ ($k \ll L_m$) prototypes for the $m$-th modality via:
\begin{equation}
    \begin{aligned}
    S^{m,*}
    &= \underset{S^m \subset X^{m}}{\operatorname{argmax}}\,
    f(S^m, X^{m}) \\
    &\text{s.t.}\quad |S^m| = k .
    \end{aligned}
    \label{equ3}
\end{equation}
which is solved with a classical greedy algorithm~\cite{jungnickel1999greedy} (detailed procedure in Appendix~\ref{app:gps_algorithm}). Intuitively, GPS selects $k$ representative tokens whose affinity coverage is high for the full modality-specific token set. 
To enable end-to-end training of the upstream visual features that shape $\hat{A}$, we route gradients through a straight-through estimator. The detailed greedy procedure and straight-through estimator formulation are provided in Appendix~\ref{app:gps_algorithm} and ~\ref{alg:ste}. 
To preserve positional information, we add a learnable embedding to each selected prototype: 
\begin{equation}
    H^{m} = S^{m,*} + \mathrm{Embed}(pos)
\end{equation}
where $pos$ is the original token indices of the selected tokens within $X^{m}$.

\subsection{Change-aware Gated Attention}
\label{sec:calgr}
GPS provides a compact visual initialization in a question-agnostic manner, but it does not decide which evidence is relevant to the current question. To perform \textit{question-conditioned} evidence selection, we further maintain a multi-level feature bank
$\mathcal{B}=\{V^{\ell}\}_{\ell=1}^{L_v}$, where $V^{\ell}$ denotes the visual embeddings from the $\ell$-th layer of the vision encoder and $L_v$ is a subset of the whole vision encoder. 
During decoding, CaGA uses language-conditioned hidden states to retrieve complementary visual evidence from $\mathcal{B}$. Specifically, at the $t$-th layer in the decoder, we compute a \textit{change-aware gated attention} module from the layer-to-layer update. 

Let $H_t$ denote the hidden states after the $t$-th decoder layer, where $H_1 = \text{concat}[H^{m};T|m=1,...,M]$, \textit{i.e.}, the first layer of the decoder takes the concatenation of GPS-initialized prototypes from all modalities and the question tokens $T$ as input. Then,
we measure the mean cosine similarity between consecutive layers 
\begin{equation}
\mathrm{sim}_t=\frac{1}{N}\sum_{n=1}^{N}\cos\!\left(\mathbf{h}_{t,n},\,\mathbf{h}_{t-1,n}\right),
\end{equation}
where $\mathbf{h}_{t,n}$ is the $n$-th token in $H_t$, and $N$ is the number of tokens. 
To inject multi-level cues only when needed and to avoid  always-on retrieval, we map this similarity to a scalar gate defined as
\begin{equation}
g_t=\sigma\!\left((\mathrm{sim}_t-\tau)\cdot s\right),
\end{equation}
where $\tau$ is a threshold and $s$ controls the sharpness, $\sigma$ denotes the sigmoid function. Finally, we update the visual token states by gated fusion:
\begin{align}
H_t &\leftarrow (1-g_t)\,H_t + g_t\,\hat{H}_t, \\
\hat{H}_t&=\mathrm{CrossAttn}(H_t, V_t),
\end{align}
where $V_t$ is the bank feature used at layer $t$.

\subsection{Marginal-Utility Regularization}
\label{sec:scrmu}
Layer-wise retrieval can collapse to trivial always-on or always-off behavior without explicit supervision. Our goal is to train the gate to activate retrieval only when it improves task performance. To achieve this, we run a retrieval-enabled pass with task loss $\mathcal{L}_{\mathrm{full}}$ and a retrieval-disabled pass (setting $g_t\rightarrow0$) with loss $\mathcal{L}_{\mathrm{dis}}$. The disabled branch is treated as a stop-gradient baseline. 

To penalize cases where retrieval fails to improve over the disabled baseline, we designed Marginal-utility loss with a soft margin $\delta$:
\begin{equation}
\mathcal{L}_{\mathrm{mu}}=
\operatorname{softplus}(
\mathcal{L}_{\mathrm{full}}
-\operatorname{stopgrad}(\mathcal{L}_{\mathrm{dis}})
+\delta ).
\end{equation}
This term is near zero when retrieval yields a lower loss than the baseline by at least $\delta$, and increases otherwise. 
The final objective is:
\begin{equation}
\mathcal{L}
=\mathcal{L}_{\mathrm{full}}+\beta\,\mathcal{L}_{\mathrm{mu}}+\mathcal{R},
\end{equation}
where $\mathcal{R}=1\times10^{-4}$ is a small numerical-stability constant to prevent zero-loss solutions.
 \section{Experiment}

\subsection{Experimental Setup}
We report multiple-choice accuracy and free-text BLEU~\cite{papineni-etal-2002-BLEU}, ROUGE-L~\cite{lin2004ROUGE}, and BERTScore~\cite{zhang2019BERTScore} across all benchmarks.
Baselines span three categories: general MLLMs (Qwen3-VL~\cite{bai2025qwen3vl}, Qwen2.5-VL~\cite{bai2025qwen25vl}), large medical VLMs (HuLu-Med~\cite{jiang2025hulumed}, Lingshu~\cite{xu2025lingshu}, OmniV~\cite{jiang2025omnivmed}), and 3D-specific models (M3D~\cite{bai2024m3d}, Merlin~\cite{blankemeier_kumar2026merlin}, RadFM, CT-CHAT), evaluated in zero-shot and fine-tuned settings.
Implementation details are in Appendix~\ref{app:implementation_details}.

\subsection{Main Results on \dataset}
\label{sec:main_results}

\begin{table*}[t]
\centering
\setlength{\tabcolsep}{1.6pt}
\resizebox{0.98\textwidth}{!}{
\begin{tabular}{p{2.3cm}|c|cccc|cccccc}
\toprule
\multirow{3}{*}{Task} & \multirow{3}{*}{Metric} & \multicolumn{4}{c|}{Zero-shot} & \multicolumn{6}{c}{Fine-tune} \\ \cline{3-12}
 &  & Qwen3-VL & Hulu-Med & Hulu-Med & Lingshu & M3D-Phi3 & M3D-L2 & M3D-L3 & Lingshu & Qwen3-VL & \model \\ \cline{3-12}
 &  & 4B & 4B & 7B & 32B & 4B & 7B & 8B & 7B & 4B & 4B \\ \midrule
\multirow{4}{2.3cm}{Global Background Assessment} & Accuracy & 23.20 & 57.25 & 59.95 & 69.43 & 76.78 & 72.83 & 66.53 & 88.23 & \underline{89.43} & \cellcolor{blue!3}\textbf{89.82} \\
 & BLEU & 72.76 & 80.34 & 82.91 & 85.12 & 91.23 & 92.87 & 90.12 & 95.92 & \underline{96.55} & \cellcolor{blue!3}\textbf{96.63} \\
 & ROUGE & 72.63 & 84.76 & 87.42 & 91.07 & 93.45 & 94.81 & 92.15 & 97.42 & \textbf{97.86} & \cellcolor{blue!3}\underline{97.85} \\
 & BERT. & 94.41 & 97.83 & 98.36 & 99.27 & 98.94 & 99.22 & 98.83 & 99.62 & \underline{99.63} & \cellcolor{blue!3}\textbf{99.68} \\ \hline
\multirow{4}{2.3cm}{Lesion Detection \& Localization} & Accuracy & 13.64 & 37.44 & 47.15 & 70.10 & 62.68 & 64.44 & 61.34 & \underline{77.12} & 73.55 & \cellcolor{blue!3}\textbf{78.48} \\
 & BLEU & 60.99 & 74.28 & 79.61 & 86.99 & 93.12 & 94.25 & 92.34 & 98.30 & \textbf{98.74} & \cellcolor{blue!3}\underline{98.61} \\
 & ROUGE & 53.11 & 70.35 & 77.84 & 89.31 & 90.22 & 91.53 & 89.64 & 92.81 & \underline{93.56} & \cellcolor{blue!3}\textbf{93.96} \\
 & BERT. & 91.20 & 95.96 & 97.12 & 98.88 & 99.48 & 99.65 & 99.41 & 99.80 & \textbf{99.88} & \cellcolor{blue!3}\underline{99.87} \\ \hline
\multirow{4}{2.3cm}{Morphological Characterization} & Accuracy & 44.79 & 18.33 & 22.36 & 30.20 & 38.78 & 35.35 & 33.24 & 55.23 & \textbf{57.68} & \cellcolor{blue!3}\underline{57.45} \\
 & BLEU & 41.22 & 35.84 & 39.72 & 54.65 & 62.15 & 64.83 & 60.31 & \underline{71.29} & 70.34 & \cellcolor{blue!3}\textbf{73.46} \\
 & ROUGE & 47.23 & 42.17 & 45.68 & 55.87 & 64.23 & 66.87 & 62.34 & 71.90 & \underline{71.92} & \cellcolor{blue!3}\textbf{72.63} \\
 & BERT. & 91.80 & 90.86 & 92.14 & 96.29 & 96.12 & 96.48 & 95.93 & 97.12 & \underline{97.24} & \cellcolor{blue!3}\textbf{97.46} \\ \hline
\multirow{4}{2.3cm}{Multi-Modal Functional Reasoning} & Accuracy & 56.01 & 57.46 & 57.46 & 50.91 & 59.32 & 72.24 & 63.21 & 78.92 & \underline{79.81} & \cellcolor{blue!3}\textbf{80.11} \\
 & BLEU & 58.80 & 68.74 & 71.36 & 81.88 & 88.12 & 90.23 & 87.34 & 94.13 & \underline{97.01} & \cellcolor{blue!3}\textbf{97.28} \\
 & ROUGE & 58.09 & 70.12 & 72.88 & 82.78 & 89.45 & 91.14 & 88.21 & 94.08 & \underline{96.97} & \cellcolor{blue!3}\textbf{97.37} \\
 & BERT. & 91.69 & 95.12 & 95.84 & 98.06 & 98.12 & 98.41 & 97.95 & 98.72 & \underline{99.35} & \cellcolor{blue!3}\textbf{99.43} \\ \hline
\multirow{4}{2.3cm}{Local Invasion \& Nodal Assessment} & Accuracy & 9.67 & 66.29 & 66.29 & 60.64 & 54.81 & 54.32 & 44.31 & 76.23 & \textbf{77.54} & \cellcolor{blue!3}\underline{77.42} \\
 & BLEU & 54.44 & 78.45 & 80.13 & 76.82 & 82.31 & 83.54 & 80.12 & \underline{87.18} & 84.14 & \cellcolor{blue!3}\textbf{88.01} \\
 & ROUGE & 57.42 & 79.22 & 81.05 & 77.83 & 83.42 & 84.63 & 81.24 & \underline{86.72} & 85.92 & \cellcolor{blue!3}\textbf{87.83} \\
 & BERT. & 92.74 & 95.64 & 96.08 & 95.27 & 97.12 & 97.34 & 96.91 & \underline{97.58} & 97.41 & \cellcolor{blue!3}\textbf{97.77} \\ \hline
\multirow{4}{2.3cm}{Holistic Diagnostic Decision} & Accuracy & 23.51 & 28.94 & 31.33 & 23.99 & 28.96 & 30.78 & 21.52 & 34.12 & \underline{35.86} & \cellcolor{blue!3}\textbf{36.53} \\
 & BLEU & 43.93 & 69.15 & 71.42 & 72.79 & 82.34 & 84.12 & 80.15 & \underline{87.82} & 87.35 & \cellcolor{blue!3}\textbf{88.22} \\
 & ROUGE & 44.62 & 70.08 & 72.35 & 73.90 & 82.13 & 83.55 & 80.13 & 85.61 & \underline{85.70} & \cellcolor{blue!3}\textbf{85.86} \\
 & BERT. & 89.30 & 95.21 & 95.62 & 95.84 & 97.15 & 97.31 & 96.94 & 97.50 & \underline{97.53} & \cellcolor{blue!3}\textbf{97.59} \\ \hline
\multirow{4}{2.3cm}{Pathology Prediction} & Accuracy & 38.04 & 49.68 & 51.33 & 47.79 & 54.74 & 54.12 & 54.32 & \underline{72.57} & 71.29 & \cellcolor{blue!3}\textbf{74.18} \\
 & BLEU & 61.47 & 85.12 & 87.46 & 88.95 & 91.12 & 92.34 & 90.21 & 95.34 & \underline{94.89} & \cellcolor{blue!3}\textbf{95.34} \\
 & ROUGE & 62.61 & 86.34 & 88.12 & 89.96 & 92.11 & 93.45 & 91.24 & \underline{95.44} & 94.99 & \cellcolor{blue!3}\textbf{95.53} \\
 & BERT. & 92.15 & 95.74 & 96.02 & 96.18 & 98.72 & 98.89 & 98.61 & \underline{99.07} & 98.98 & \cellcolor{blue!3}\textbf{99.10} \\ \hline
\multirow{4}{2.3cm}{Average} & Accuracy & 29.84 & 45.06 & 47.98 & 50.44 & 53.72 & 54.87 & 49.21 & 68.92 & \underline{69.45} & \cellcolor{blue!3}\textbf{70.57} \\
 & BLEU & 56.23 & 70.27 & 73.23 & 78.17 & 84.34 & 86.03 & 82.94 & 90.00 & \underline{89.87} & \cellcolor{blue!3}\textbf{91.08} \\
 & ROUGE & 56.53 & 71.86 & 75.05 & 80.10 & 85.00 & 86.57 & 83.56 & 89.14 & \underline{89.56} & \cellcolor{blue!3}\textbf{90.15} \\
 & BERT. & 91.90 & 95.19 & 95.88 & 97.11 & 97.95 & 98.19 & 97.80 & 98.49 & \underline{98.58} & \cellcolor{blue!3}\textbf{98.70} \\ \bottomrule
\end{tabular}
}
\vspace{-0.5em}
\caption{Main performance on \dataset.
\textbf{Bold} and \underline{underline} indicate the best and second-best performance.}\label{tab:mri_s1_main}
\vspace{-0.7em}
\end{table*}

\begin{wraptable}{r}{0.55\textwidth}
\vspace{-1.2\baselineskip}
\centering
\setlength{\tabcolsep}{2.0pt}
\renewcommand{\arraystretch}{0.88}
\resizebox{0.96\linewidth}{!}{
\begin{tabular}{lcccc}
\toprule
Method & Vis. Tokens & Lat. (ms) $\downarrow$ & Accuracy & BERT. \\
\midrule
Baseline & 12544 & 952.6 & 69.18 & 98.42 \\
\midrule
VisionZip & 256 & 630.8 & 67.63 & 98.16 \\
VisionZip & 512 & 698.4 & 68.72 & 98.34 \\
VisionZip & 1024 & 829.6 & 69.03 & 98.39 \\
\midrule
DivPrune & 256 & 605.7 & 67.21 & 98.08 \\
DivPrune & 512 & 625.6 & 68.24 & 98.25 \\
DivPrune & 1024 & 863.3 & 68.61 & 98.31 \\
\midrule
MMToK & 256 & 605.6 & 66.94 & 98.03 \\
MMToK & 512 & 618.8 & 67.86 & 98.18 \\
MMToK & 1024 & 829.4 & 68.37 & 98.27 \\
\midrule
\rowcolor{blue!3}
\model & 256 & 598.4 & 69.73 & 98.48 \\
\rowcolor{blue!3}
\model & 512 & 644.1 & 70.21 & 98.60 \\
\rowcolor{blue!3}
\model & 1024 & 833.2 & 69.88 & 98.52 \\
\bottomrule
\end{tabular}
}
\vspace{-0.5em}
\caption{Latency--performance comparison with visual token pruning methods on \dataset. }\label{tab:pruning_efficiency_bremris}
\vspace{-0.5\baselineskip}
\end{wraptable}

Table~\ref{tab:mri_s1_main} reports results across all seven workflow tasks.
Zero-shot models, including domain-pretrained medical specialists, remain substantially below fine-tuned models, with larger gaps on tasks involving spatial localization and nodal reasoning.
Among fine-tuned models, 3D-specific architectures (M3D) and Lingshu show lower performance than general transformers, as M3D is pre-trained on single-modality CT data and Lingshu is not specifically optimized for multi-sequence volumetric inputs.
\model at 4B scale achieves the best overall average with the largest improvements on cross-sequence integration tasks, while on Morphological Characterization, multiple-choice accuracy marginally trails fine-tuned Qwen3-VL, suggesting compact prototypes may not fully preserve fine-grained shape information.
\subsection{Efficiency Analysis}
\label{sec:efficiency}

Table~\ref{tab:pruning_efficiency_bremris} compares \model against VisionZip~\cite{yang2025visionzip}, DivPrune~\cite{alvar2025divprune}, and MMToK~\cite{dong2026mmtok} under the same backbone (Qwen2.5-VL~\cite{bai2025qwen25vl}) and budgets.
Pruning methods fall below the full-token baseline in accuracy at all budgets, since irreversible token removal discards visual content that cannot be recovered. Although \model is not always the fastest, its latency remains within a practical range.
\model at $k{=}512$ achieves 70.21\% accuracy in 644\,ms, exceeding the full-token baseline (69.18\%, 953\,ms) in both accuracy and speed. Baseline implementation details are in Appendix~\ref{app:implementation_details}.

\WFclear
\subsection{Modality Robustness and Interpretability}
\label{sec:robustness}

\begin{wrapfigure}{r}{0.56\textwidth}
    \vspace{-0.7\baselineskip}
    \centering
    \includegraphics[width=\linewidth]{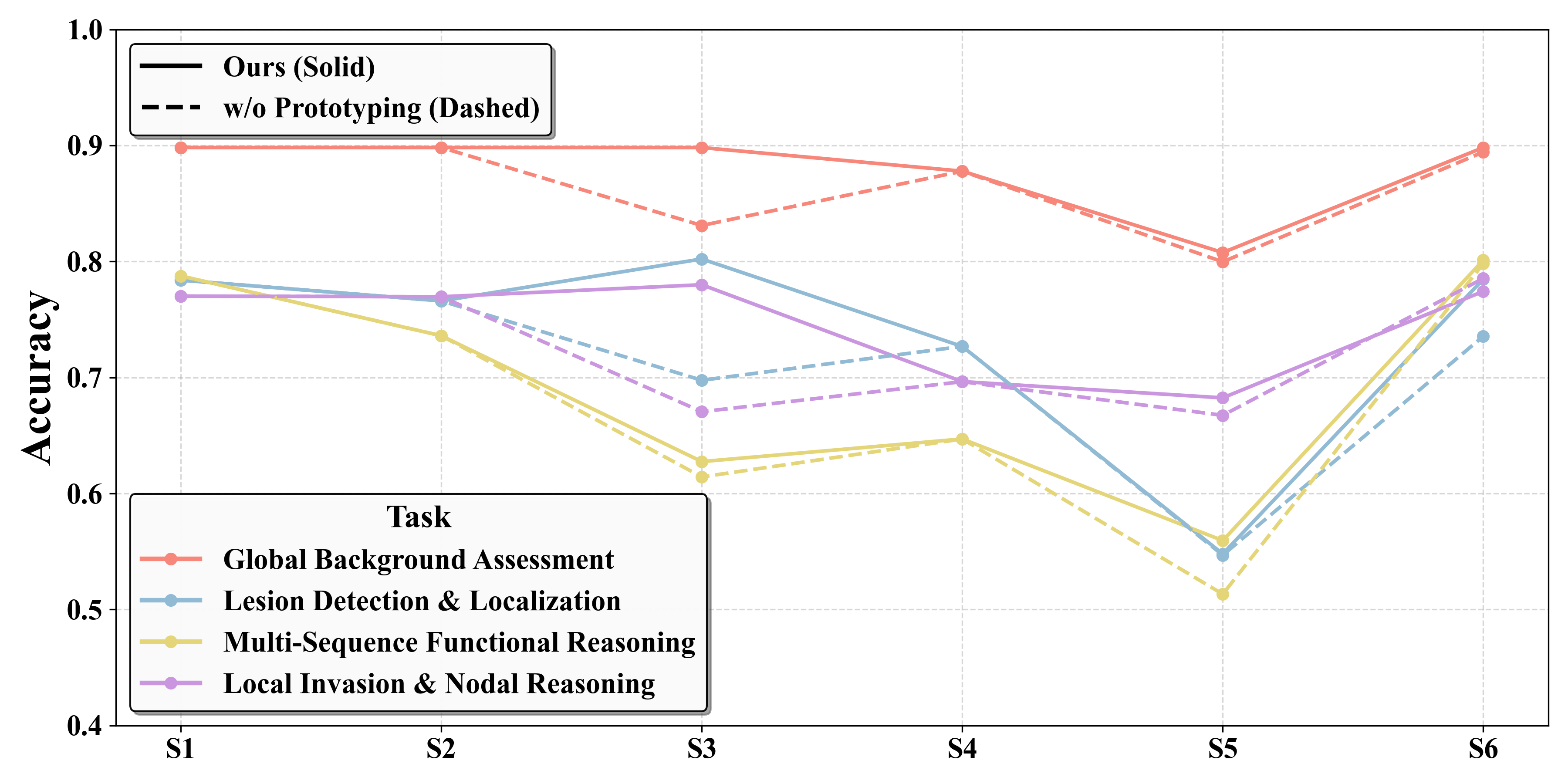}
    \vspace{-0.5em}
    \caption{\textbf{Missing-modality robustness.}
    Multiple-choice accuracy under six modality settings (S1--S6).}
    \label{fig:modality_ablation}
    \vspace{-0.5\baselineskip}
\end{wrapfigure}

\textbf{Missing-modality robustness.}
We compare \model against its variant without GPS prototype selection (w/o Proto.) across six modality settings (S1--S6; Figure~\ref{fig:modality_ablation}).
Both models perform comparably under S1, S2, and S4, which retain primary structural and enhancement sequences, but degrade more substantially under S3 (DWI, ADC, and T2w, without DCE or T1w) and S5 (without T1w), where key reference modalities are absent.
\model maintains a smaller accuracy drop relative to S6 in these settings, as CaGA redistributes retrieval weights across available sequences rather than relying on fixed token positions.
Details and per-task free-text results are in Appendix~\ref{app:add_modality_setting}.

\begin{figure*}[!t]
    \centering
    \begin{subfigure}{\linewidth}
        \centering
        \includegraphics[width=0.98\linewidth]{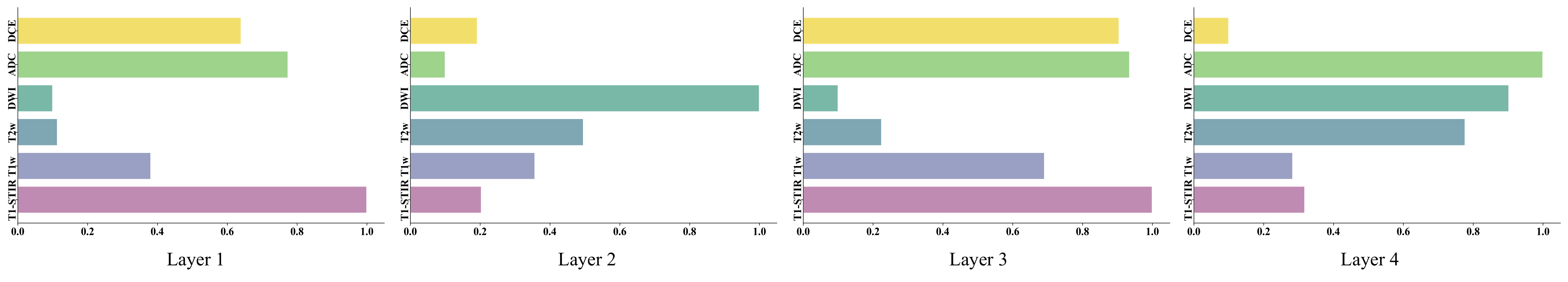}
        \vspace{-0.5em}
        \caption{Layer-wise modality contribution: mean gated cross-attention weight per modality across decoder transformer layers.}
        \label{fig:importance}
        \vspace{-0.2em}
    \end{subfigure}
    \begin{subfigure}{\linewidth}
        \centering
        \includegraphics[width=0.98\linewidth]{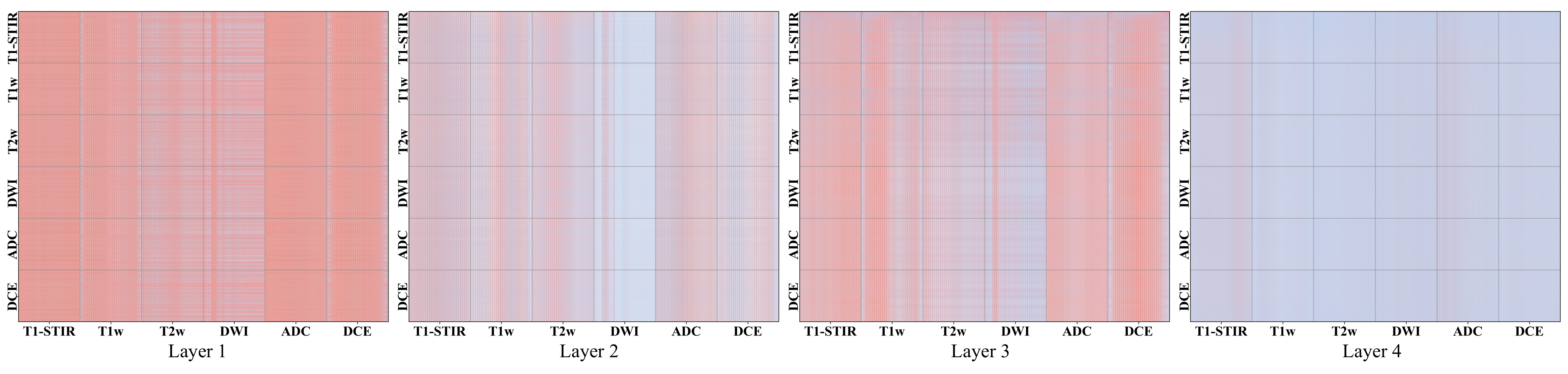}
        \vspace{-0.5em}
        \caption{Token-level gated attention maps across decoder layers: attention sparsifies with depth as sufficient evidence is integrated.}
        \label{fig:attnmap}
        \vspace{-0.5em}
    \end{subfigure}
    \vspace{-1.5em}
    \caption{Interpretability visualization of \model on the question \textit{``What is the delayed phase enhancement kinetics of the mass in the upper inner quadrant of the right breast?''} with six MRI modalities.}
    \label{fig:interpretability}
    \vspace{-1em}
\end{figure*}

\noindent\textbf{Layer-wise modality preference.}
Fig.~\ref{fig:importance} shows that early layers broadly attend to structural modalities (e.g., T1-STIR) for anatomical perception and lesion localization. In intermediate layers, the model increasingly focuses on functionally informative modalities (DCE, DWI, ADC), indicating active question-aware functional reasoning. 

\noindent\textbf{Gated attention sparsification.}
Fig.~\ref{fig:attnmap} illustrates the progressive sparsification of token-level gated attention across transformer layers. Early layers exhibit dense cross-modal attention for global anatomical exploration, while intermediate layers progressively suppress redundant tokens and focus on diagnostically relevant regions. In deeper layers, attention becomes highly sparse, indicating that the model has distilled clinically relevant information into compact semantic representations. This hierarchical transition from broad exploration to selective semantic condensation reflects effective question-aware clinical reasoning.

\WFclear

\begin{table*}[!t]
\centering
\setlength{\tabcolsep}{3.0pt}
\renewcommand{\arraystretch}{1.0}
\resizebox{0.98\textwidth}{!}{\begin{tabular}{>{\raggedright\arraybackslash}p{2.8cm}|c|ccccccc|cccc}
\toprule    
\multirow{3}{*}{Task} & \multirow{3}{*}{Metric} & \multicolumn{7}{c|}{Zero-shot} & \multicolumn{4}{c}{Finetune} \\ \cline{3-13}
 &  & Q2.5VL & Q3VL & RadFM & M3D-L2 & M3D-P3 & OmniV & Lingshu & M3D-L2 & M3D-P3 & Qwen3-VL & \model \\ \cline{3-13}
 &  & 4B & 4B & 13B & 7B & 4B & 1.5B & 7B & 7B & 4B & 4B & 4B \\ \midrule
Exist.\ Det. & Accuracy & 19.52 & 3.52 & 29.20 & 18.00 & 40.25 & 28.66 & 59.60 & 81.09 & 82.43 & \underline{82.52} & \cellcolor{blue!3}\textbf{82.62} \\ \hline
Static T.\ Diag. & Accuracy & 0.00 & 0.00 & 44.11 & 25.47 & 25.40 & 22.96 & 6.02 & \underline{51.20} & 49.30 & 47.70 & \cellcolor{blue!3}\textbf{51.77} \\ \hline
Longit.\ T.\ Diag. & Accuracy & 0.29 & 0.15 & 42.99 & 24.17 & 24.31 & 24.23 & 12.13 & \underline{74.78} & 74.77 & 74.39 & \cellcolor{blue!3}\textbf{75.28} \\ \hline
\multirow{3}{2.8cm}{Medical Measurement} & BLEU & 1.78 & 0.11 & 3.34 & 15.95 & 2.55 & 2.52 & 2.50 & 30.54 & 33.52 & \underline{35.01} & \cellcolor{blue!3}\textbf{36.78} \\
 & ROUGE & 4.30 & 0.08 & 6.62 & 23.24 & 5.63 & 7.88 & 5.45 & 36.06 & 36.46 & \underline{38.49} & \cellcolor{blue!3}\textbf{39.01} \\
 & BERT. & 84.20 & 76.65 & 86.85 & 91.50 & 85.74 & 85.66 & 83.81 & 94.65 & 94.86 & \textbf{96.01} & \cellcolor{blue!3}\underline{95.93} \\ \hline
\multirow{3}{2.8cm}{Image Observation} & BLEU & 3.51 & 0.26 & 13.48 & 10.69 & 16.31 & 16.42 & 5.34 & 31.28 & 39.66 & \underline{48.08} & \cellcolor{blue!3}\textbf{48.81} \\
 & ROUGE & 8.84 & 0.44 & 19.14 & 20.82 & 23.19 & 26.69 & 12.25 & 39.12 & 50.52 & \underline{52.48} & \cellcolor{blue!3}\textbf{54.16} \\
 & BERT. & 84.53 & 76.22 & 87.16 & 86.61 & 86.92 & 88.29 & 85.67 & 90.00 & 92.19 & \underline{92.92} & \cellcolor{blue!3}\textbf{93.09} \\ \hline
\multirow{3}{2.8cm}{Anomaly Detection} & BLEU & 2.93 & 0.24 & 11.00 & 9.10 & 15.06 & 13.47 & 3.71 & 25.25 & 33.28 & \textbf{39.95} & \cellcolor{blue!3}\underline{39.83} \\
 & ROUGE & 9.17 & 0.54 & 17.62 & 18.64 & 23.19 & 25.72 & 9.46 & 33.76 & 42.45 & \underline{43.96} & \cellcolor{blue!3}\textbf{45.58} \\
 & BERT. & 84.47 & 76.65 & 86.76 & 86.07 & 87.11 & 88.21 & 84.81 & 89.16 & 90.72 & \underline{91.43} & \cellcolor{blue!3}\textbf{91.58} \\ \bottomrule
\end{tabular}
}
\vspace{-0.5em}
\caption{Comparison of zero-shot and finetuned performance across six tasks in 3D-RAD.
Exist.\ Det.: Existence Detection; Static T.\ Diag.: Static Temporal Diagnosis; Longit.\ T.\ Diag.: Longitudinal Temporal Diagnosis.
Q2.5VL: Qwen2.5-VL; Q3VL: Qwen3-VL. 
\textbf{Bold} and \underline{underline} indicate the best and second-best performance.}\label{tab:3drad_table1}
\vspace{-0.5em}
\end{table*}

\Needspace{0.58\textheight}
\subsection{Public 3D Benchmarks}
\label{sec:public_benchmarks}
\begin{wraptable}{r}{0.60\textwidth}
\vspace{-0.7\baselineskip}
\centering
\footnotesize
\setlength{\tabcolsep}{1.5pt}
\renewcommand{\arraystretch}{1.07}
\resizebox{\linewidth}{!}{
\begin{tabular}{llcccccc}
\toprule
Type & Metric & Merlin & M3D-L2 & M3D-P3 & CT-CHAT & RadFM & \model \\
\hline
\multirow{2}{*}{Meas} & MC & 0.256 & 0.733 & 0.740 & \textbf{0.740} & \underline{0.739} & \cellcolor{blue!3}0.735 \\
 & FT & 0.364 & 0.370 & 0.366 & 0.379 & \underline{0.434} & \cellcolor{blue!3}\textbf{0.521} \\
\hline
\multirow{2}{*}{Recog} & MC & 0.661 & 0.664 & 0.661 & 0.664 & \textbf{0.789} & \cellcolor{blue!3}\underline{0.707} \\
 & FT & 0.664 & 0.665 & 0.660 & 0.663 & \textbf{0.812} & \cellcolor{blue!3}\underline{0.715} \\
\hline
\multirow{2}{*}{VisRsn} & MC & 0.382 & 0.616 & 0.627 & 0.620 & \underline{0.625} & \cellcolor{blue!3}\textbf{0.673} \\
 & FT & 0.392 & 0.408 & 0.420 & 0.428 & \underline{0.439} & \cellcolor{blue!3}\textbf{0.499} \\
\hline
\multirow{2}{*}{MedRsn} & MC & 0.446 & 0.557 & 0.560 & 0.555 & \underline{0.584} & \cellcolor{blue!3}\textbf{0.656} \\
 & FT & 0.550 & 0.562 & 0.561 & 0.549 & \underline{0.629} & \cellcolor{blue!3}\textbf{0.663} \\
\hline
\multirow{2}{*}{Total} & MC & 0.440 & 0.626 & 0.632 & 0.628 & \underline{0.662} & \cellcolor{blue!3}\textbf{0.687} \\
 & FT & 0.478 & 0.489 & 0.493 & 0.497 & \underline{0.555} & \cellcolor{blue!3}\textbf{0.599} \\
\bottomrule
\end{tabular}
}
\vspace{-0.5em}
\caption{Average DeepTumorVQA performance. Meas: Measurement; Recog: Recognition; VisRsn: Visual Reasoning; MedRsn: Medical Reasoning. MC: multiple-choice accuracy; FT: average free-text score.}\label{tab:deeptumorvqa}
\vspace{-0.5\baselineskip}
\end{wraptable}
Tables~\ref{tab:3drad_table1} and~\ref{tab:deeptumorvqa} evaluate \model on 3D-RAD~\cite{gai20253drad} and DeepTumorVQA~\cite{chen2025deeptumor}, two single-volume benchmarks that serve as transfer tests for the selective exposure principle.
On 3D-RAD, \model achieves the highest accuracy on all three classification and temporal tasks and leads on the majority of generation metrics, remaining closely competitive with fine-tuned Qwen3-VL throughout.
On DeepTumorVQA, \model ranks first overall on both multiple-choice and free-text metrics and leads on Visual Reasoning and Medical Reasoning, while models with large-scale CT pretraining outperform \model on Measurement and Recognition.
Full results are in Appendix~\ref{app:deeptumorvqa_full}.

\WFclear
\subsection{Ablation Study}
\label{sec:ablation}

\begin{wraptable}{r}{0.55\textwidth}
\vspace{-0.7\baselineskip}
\centering
\setlength{\tabcolsep}{4pt}
\renewcommand{\arraystretch}{0.98}
\resizebox{\linewidth}{!}{
\begin{tabular}{lccccc}
\toprule
Setting & Acc. & BLEU & ROUGE & BERT. & $k$ \\
\midrule
Baseline Ft. & 69.45 & 89.87 & 89.56 & 98.58 & 4608 \\
w/o CaGA & 64.36 & 88.92 & 85.30 & 95.42 & 512 \\
w/o Regul. & 68.17 & 87.04 & 89.16 & 97.63 & 512 \\
Q-GPS & 69.92 & 90.21 & 89.83 & 97.66 & 512 \\
\rowcolor{blue!3}
\model & 70.57 & 91.08 & 90.15 & 98.70 & 512 \\
\bottomrule
\end{tabular}
}
\vspace{-0.5em}
\caption{Compact ablation summary on \dataset. We report average metrics across seven tasks and the number of visual prototypes exposed to the decoder.}\label{tab:compact_ablation_bremris}
\vspace{-0.5\baselineskip}
\end{wraptable}
\textbf{Component contributions.}
Table~\ref{tab:compact_ablation_bremris} isolates each design choice at a fixed 512-token budget.
Removing CaGA causes the largest drop, confirming that GPS prototypes alone cannot recover the fine-grained evidence compressed away at this budget and that the multi-level bank with on-demand retrieval is necessary.
Removing SCR-MU (\textit{w/o Regul.}) degrades free-text quality most: Fig.~\ref{fig:attnmap} shows that without the marginal-utility penalty late-layer attention remains dense rather than sparsifying, indicating that SCR-MU is what converts the gate from always-on to selective.
The question-conditioned GPS variant (\textit{Q-GPS}) underperforms greedy GPS, supporting the design choice of keeping GPS question-agnostic and delegating question conditioning to CaGA where full decoder context is available.
Full 3D-RAD ablation results are in Appendix~\ref{abl:3drad}, and hyperparameter sensitivity analyses are in Appendix~\ref{app:abl_all}.

\WFclear
\FloatBarrier
\begin{figure}[!t]
    \centering
    \includegraphics[width=0.68\linewidth]{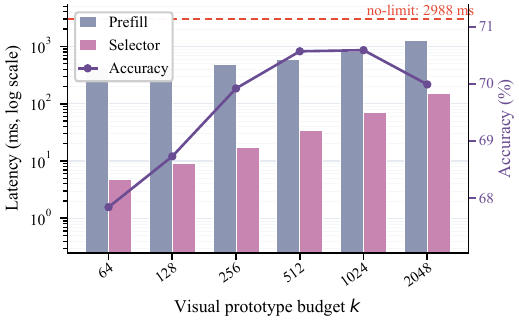}
    \vspace{-0.5em}
    \caption{\textbf{Prototype-budget latency profile.}
    Latency and performance under prototype budgets $k$, with the full-token no-limit setting shown as a dashed reference.}
    \label{fig:k_sweep_latency}
    \vspace{-1em}
\end{figure}

\noindent\textbf{Prototype budget.}
Fig.~\ref{fig:k_sweep_latency} shows that increasing $k$ improves accuracy up to $k{=}512$, beyond which performance plateaus or declines as redundant tokens dilute the prototype set while latency continues to rise.
We therefore set $k{=}512$ as the default, where the accuracy gain over smaller budgets is substantial and latency remains well below the full-token baseline.
More results are in Appendix~\ref{abl:3drad}.

\noindent\textbf{Backbone compatibility.}
We apply \model to two distinct backbone families: Qwen2.5-VL-3B (70.21\% Acc) and Qwen3-VL at 4B and 8B (70.57\% and 72.13\%).
Both families yield consistent gains over their respective fine-tuned baselines, demonstrating that the selective exposure design is not specific to any single backbone and can be integrated without architecture-level modifications.
Hyperparameter sensitivity for $\tau_v$, $k$, $s$, and $\beta$ and full task-wise results are in Appendix~\ref{app:cross-model-scaling}.
 
\section{Conclusion}
We introduce \textbf{\dataset}, a clinically curated benchmark with 1.19M QA pairs from 12.9K multi-sequence breast MRI cases, and \textbf{\model}, a framework for selective visual exposure and dynamic evidence retrieval in volumetric medical VQA.
\model achieves compact global coverage of multi-sequence volumes through prototype selection and retrieves fine-grained feature evidence during decoding, with a self-consistency objective that suppresses redundant retrieval.
On \dataset, \model outperforms same-scale fine-tuned baselines, with the largest gains on cross-sequence integration tasks and a more favorable accuracy-latency tradeoff than token pruning methods.
Results on public 3D single-volume benchmarks further confirm selective visual exposure as a broadly applicable principle for volumetric medical reasoning.

\section*{Limitations}
Despite the encouraging results, this work has several limitations that point to directions for future research. Firstly, the free-text generation tasks in \dataset are evaluated with automated language metrics, which may not fully capture clinical nuance, and radiologist-grounded human evaluation is left for future work. Secondly, the GPS module currently uses a fixed prototype budget per sequence, and adapting compression ratios to sequence-level complexity remains unexplored. Finally, while \model is validated on algorithmic benchmarks, integration studies within real-world radiology reading workflows are beyond the scope of this paper. Addressing these limitations will further strengthen the clinical applicability and practical deployment of the proposed framework.

\section*{Acknowledgements}  
This research was partially supported by National Natural Science Foundation of China under Grant No. 62476246, "Pioneer" and "Leading Goose" R\&D Program of Zhejiang under Grant No. 2025C02120, and GuangZhou City’s Key R\&D Program of China under Grant No. 2024B01J1301. This work was also partially supported by DAMO Academy through DAMO Academy Research Intern Program. Jiajin Zhang was partially supported by the Zhejiang Province Postdoctoral Research Excellence Funding Program under Grant No. ZJ2025048.

\bibliographystyle{assets/plainnat}
\bibliography{custom}

\begin{thebibliography}{32}
\providecommand{\natexlab}[1]{#1}
\providecommand{\url}[1]{\texttt{#1}}
\expandafter\ifx\csname urlstyle\endcsname\relax
  \providecommand{\doi}[1]{doi: #1}\else
  \providecommand{\doi}{doi: \begingroup \urlstyle{rm}\Url}\fi

\bibitem[Alayrac et~al.(2022)Alayrac, Donahue, Luc, Miech, Barr, Hasson, Lenc, Mensch, Millican, Reynolds, et~al.]{alayrac2022flamingo}
Jean-Baptiste Alayrac, Jeff Donahue, Pauline Luc, Antoine Miech, Iain Barr, Yana Hasson, Karel Lenc, Arthur Mensch, Katherine Millican, Malcolm Reynolds, et~al.
\newblock Flamingo: a visual language model for few-shot learning.
\newblock In \emph{Advances in Neural Information Processing Systems}, 2022.

\bibitem[Alvar et~al.(2025)Alvar, Singh, Akbari, and Zhang]{alvar2025divprune}
Saeed~Ranjbar Alvar, Gursimran Singh, Mohammad Akbari, and Yong Zhang.
\newblock Divprune: Diversity-based visual token pruning for large multimodal models.
\newblock In \emph{IEEE/CVF Conference on Computer Vision and Pattern Recognition (CVPR)}, 2025.

\bibitem[Bai et~al.(2024)Bai, Du, Huang, Meng, and Zhao]{bai2024m3d}
Fan Bai, Yuxin Du, Tiejun Huang, Max Q-H Meng, and Bo~Zhao.
\newblock M3d: Advancing 3d medical image analysis with multi-modal large language models.
\newblock \emph{arXiv preprint arXiv:2404.00578}, 2024.

\bibitem[Bai et~al.(2025{\natexlab{a}})Bai, Cai, Chen, Chen, Chen, Cheng, Deng, Ding, Gao, Ge, et~al.]{bai2025qwen3vl}
Shuai Bai, Yuxuan Cai, Ruizhe Chen, Keqin Chen, Xionghui Chen, Zesen Cheng, Lianghao Deng, Wei Ding, Chang Gao, Chunjiang Ge, et~al.
\newblock Qwen3-vl technical report.
\newblock \emph{arXiv preprint arXiv:2511.21631}, 2025{\natexlab{a}}.

\bibitem[Bai et~al.(2025{\natexlab{b}})Bai, Chen, Liu, Wang, Ge, Song, Dang, Wang, Wang, Tang, et~al.]{bai2025qwen25vl}
Shuai Bai, Keqin Chen, Xuejing Liu, Jialin Wang, Wenbin Ge, Sibo Song, Kai Dang, Peng Wang, Shijie Wang, Jun Tang, et~al.
\newblock Qwen2.5-vl technical report.
\newblock \emph{arXiv preprint arXiv:2502.13923}, 2025{\natexlab{b}}.

\bibitem[Blankemeier et~al.(2026)Blankemeier, Kumar, Cohen, Liu, Liu, Van~Veen, Gardezi, Yu, Paschali, Chen, Delbrouck, Reis, Holland, Truyts, Bluethgen, Wu, Lian, Jensen, Ostmeier, Varma, Valanarasu, Fang, Huo, Nabulsi, Ardila, Weng, Amaro~Junior, Ahuja, Fries, Shah, Zaharchuk, Willis, Yala, Johnston, Boutin, Wentland, Langlotz, Hom, Gatidis, and Chaudhari]{blankemeier_kumar2026merlin}
Louis Blankemeier, Ashwin Kumar, Joseph~Paul Cohen, Jiaming Liu, Longchao Liu, Dave Van~Veen, Syed Jamal~Safdar Gardezi, Hongkun Yu, Magdalini Paschali, Zhihong Chen, Jean-Benoit Delbrouck, Eduardo Reis, Robbie Holland, Cesar Truyts, Christian Bluethgen, Yufu Wu, Long Lian, Malte Engmann~Kjeldskov Jensen, Sophie Ostmeier, Maya Varma, Jeya Maria~Jose Valanarasu, Zhongnan Fang, Zepeng Huo, Zaid Nabulsi, Diego Ardila, Wei-Hung Weng, Edson Amaro~Junior, Neera Ahuja, Jason Fries, Nigam~H. Shah, Greg Zaharchuk, Marc Willis, Adam Yala, Andrew Johnston, Robert~D. Boutin, Andrew Wentland, Curtis~P. Langlotz, Jason Hom, Sergios Gatidis, and Akshay~S. Chaudhari.
\newblock Merlin: a computed tomography vision-language foundation model and dataset.
\newblock \emph{Nature}, 2026.
\newblock \doi{10.1038/s41586-026-10181-8}.
\newblock \url{https://doi.org/10.1038/s41586-026-10181-8}.

\bibitem[Bolya et~al.(2023)Bolya, Fu, Dai, Zhang, Feichtenhofer, and Hoffman]{bolya2023token}
Daniel Bolya, Cheng-Yang Fu, Xiaoliang Dai, Peizhao Zhang, Christoph Feichtenhofer, and Judy Hoffman.
\newblock Token merging: Your vit but faster.
\newblock In \emph{The Eleventh International Conference on Learning Representations}, 2023.
\newblock \url{https://openreview.net/forum?id=JroZRaRw7Eu}.

\bibitem[Chen et~al.(2024)Chen, Zhao, Liu, Bai, Lin, Zhou, and Chang]{Chen2024fastv}
Liang Chen, Haozhe Zhao, Tianyu Liu, Shuai Bai, Junyang Lin, Chang Zhou, and Baobao Chang.
\newblock An image is worth 1/2 tokens after layer 2: Plug-and-play inference acceleration for large vision-language models.
\newblock In \emph{Computer Vision – ECCV 2024: 18th European Conference, Milan, Italy, September 29–October 4, 2024, Proceedings, Part LXXXI}, page 19–35, Berlin, Heidelberg, 2024. Springer-Verlag.
\newblock ISBN 978-3-031-73003-0.
\newblock \doi{10.1007/978-3-031-73004-7_2}.
\newblock \url{https://doi.org/10.1007/978-3-031-73004-7_2}.

\bibitem[Chen et~al.(2025)Chen, Xiao, Bassi, Zhou, Er, Hamamci, Zhou, and Yuille]{chen2025deeptumor}
Yixiong Chen, Wenjie Xiao, Pedro~RAS Bassi, Xinze Zhou, Sezgin Er, Ibrahim~Ethem Hamamci, Zongwei Zhou, and Alan Yuille.
\newblock Are vision language models ready for clinical diagnosis? a 3d medical benchmark for tumor-centric visual question answering.
\newblock \emph{arXiv preprint arXiv:2505.18915}, 2025.

\bibitem[Dong et~al.(2026)Dong, Hu, Zhang, Yin, Fu, and Qian]{dong2026mmtok}
Sixun Dong, Juhua Hu, Mian Zhang, Ming Yin, Yanjie Fu, and Qi~Qian.
\newblock {MMT}ok: Multimodal coverage maximization for efficient inference of {VLM}s.
\newblock In \emph{ICLR}, 2026.

\bibitem[Fang et~al.(2026)Fang, Guo, Jiang, He, Li, and Xu]{fang2026photon}
Chengyu Fang, Heng Guo, Zheng Jiang, Chunming He, Xiu Li, and Minfeng Xu.
\newblock Photon: Speedup volume understanding with efficient multimodal large language models.
\newblock In \emph{The Fourteenth International Conference on Learning Representations}, 2026.
\newblock \url{https://openreview.net/forum?id=xsSJw6jJBL}.

\bibitem[Gai et~al.(2025)Gai, Liu, Li, Meng, Wu, and Liu]{gai20253drad}
Xiaotang Gai, Jiaxiang Liu, Yichen Li, Zijie Meng, Jian Wu, and Zuozhu Liu.
\newblock 3d-{RAD}: A comprehensive 3d radiology med-{VQA} dataset with multi-temporal analysis and diverse diagnostic tasks.
\newblock In \emph{The Thirty-ninth Annual Conference on Neural Information Processing Systems Datasets and Benchmarks Track}, 2025.
\newblock \url{https://openreview.net/forum?id=VB2cgrlikN}.

\bibitem[Hamamci et~al.(2024)Hamamci, Er, and Menze]{hamamci2024ct2rep}
Ibrahim~Ethem Hamamci, Sezgin Er, and Bjoern Menze.
\newblock Ct2rep: Automated radiology report generation for 3d medical imaging.
\newblock In \emph{International Conference on Medical Image Computing and Computer-Assisted Intervention}, pages 476--486. Springer, 2024.

\bibitem[Hamamci et~al.(2026)Hamamci, Er, Wang, Almas, Simsek, Esirgun, Dogan, Durugol, Hou, Shit, et~al.]{hamamci2026ct-rate}
Ibrahim~Ethem Hamamci, Sezgin Er, Chenyu Wang, Furkan Almas, Ayse~Gulnihan Simsek, Sevval~Nil Esirgun, Irem Dogan, Omer~Faruk Durugol, Benjamin Hou, Suprosanna Shit, et~al.
\newblock Generalist foundation models from a multimodal dataset for 3d computed tomography.
\newblock \emph{Nature Biomedical Engineering}, pages 1--19, 2026.

\bibitem[Hu et~al.(2024)Hu, Li, Lu, Shao, He, Qiao, and Luo]{hu2024omnimedvqa}
Yutao Hu, Tianbin Li, Quanfeng Lu, Wenqi Shao, Junjun He, Yu~Qiao, and Ping Luo.
\newblock Omnimedvqa: A new large-scale comprehensive evaluation benchmark for medical lvlm.
\newblock In \emph{Proceedings of the IEEE/CVF Conference on Computer Vision and Pattern Recognition}, pages 22170--22183, 2024.

\bibitem[Jiang et~al.(2025{\natexlab{a}})Jiang, Wang, Song, Hu, Zhou, Pu, Zhang, Yang, Feng, Zhou, et~al.]{jiang2025hulumed}
Songtao Jiang, Yuan Wang, Sibo Song, Tianxiang Hu, Chenyi Zhou, Bin Pu, Yan Zhang, Zhibo Yang, Yang Feng, Joey~Tianyi Zhou, et~al.
\newblock Hulu-med: A transparent generalist model towards holistic medical vision-language understanding.
\newblock \emph{arXiv preprint arXiv:2510.08668}, 2025{\natexlab{a}}.

\bibitem[Jiang et~al.(2025{\natexlab{b}})Jiang, Wang, Song, Zhang, Meng, Lei, Wu, Sun, and Liu]{jiang2025omnivmed}
Songtao Jiang, Yuan Wang, Sibo Song, Yan Zhang, Zijie Meng, Bohan Lei, Jian Wu, Jimeng Sun, and Zuozhu Liu.
\newblock Omniv-med: Scaling medical vision-language model for universal visual understanding.
\newblock \emph{arXiv preprint arXiv:2504.14692}, 2025{\natexlab{b}}.

\bibitem[Jungnickel(1999)]{jungnickel1999greedy}
Dieter Jungnickel.
\newblock The greedy algorithm.
\newblock In \emph{Graphs, networks and algorithms}. 1999.

\bibitem[Li et~al.(2023)Li, Wong, Zhang, Usuyama, Liu, Yang, Naumann, Poon, and Gao]{li2023llavamed}
Chunyuan Li, Cliff Wong, Sheng Zhang, Naoto Usuyama, Haotian Liu, Jianwei Yang, Tristan Naumann, Hoifung Poon, and Jianfeng Gao.
\newblock {LL}a{VA}-med: Training a large language-and-vision assistant for biomedicine in one day.
\newblock In \emph{Thirty-seventh Conference on Neural Information Processing Systems Datasets and Benchmarks Track}, 2023.
\newblock \url{https://openreview.net/forum?id=GSuP99u2kR}.

\bibitem[Lin(2004)]{lin2004ROUGE}
Chin-Yew Lin.
\newblock Rouge: A package for automatic evaluation of summaries.
\newblock In \emph{Text summarization branches out}, pages 74--81, 2004.

\bibitem[Lin et~al.(2025)Lin, Zhang, Li, Yuan, Yu, Li, He, Jiang, Li, xiaohui, Tang, Xiao, Lin, Zhuang, and Ooi]{lin2025healthgpt}
Tianwei Lin, Wenqiao Zhang, Sijing Li, Yuqian Yuan, Binhe Yu, Haoyuan Li, Wanggui He, Hao Jiang, Mengze Li, Song xiaohui, Siliang Tang, Jun Xiao, Hui Lin, Yueting Zhuang, and Beng~Chin Ooi.
\newblock Health{GPT}: A medical large vision-language model for unifying comprehension and generation via heterogeneous knowledge adaptation.
\newblock In \emph{Forty-second International Conference on Machine Learning}, 2025.
\newblock \url{https://openreview.net/forum?id=WbP2OwMULq}.

\bibitem[Lin et~al.(2026)Lin, Qiu, Zhang, Liu, Xie, Gao, Fan, Li, Li, Xie, LU, Zhuang, Xia, Zhang, and Ooi]{lin2026omnict}
Tianwei Lin, Zhongwei Qiu, Wenqiao Zhang, Jiang Liu, Yihan Xie, Mingjian Gao, Zhenxuan Fan, Zhaocheng Li, Sijing Li, Zhongle Xie, Peng LU, Yueting Zhuang, Yingda Xia, Ling Zhang, and Beng~Chin Ooi.
\newblock Omni{CT}: Towards a unified slice-volume {LVLM} for comprehensive {CT} analysis.
\newblock In \emph{The Fourteenth International Conference on Learning Representations}, 2026.
\newblock \url{https://openreview.net/forum?id=nrZI64gTvC}.

\bibitem[Minhas and Oliver(2022)]{mri_basics}
Atul~Singh Minhas and Ruth Oliver.
\newblock Magnetic resonance imaging basics.
\newblock \emph{Electrical properties of tissues: quantitative magnetic resonance mapping}, pages 47--82, 2022.

\bibitem[Papineni et~al.(2002)Papineni, Roukos, Ward, and Zhu]{papineni-etal-2002-BLEU}
Kishore Papineni, Salim Roukos, Todd Ward, and Wei-Jing Zhu.
\newblock {B}leu: a method for automatic evaluation of machine translation.
\newblock In Pierre Isabelle, Eugene Charniak, and Dekang Lin, editors, \emph{Proceedings of the 40th Annual Meeting of the Association for Computational Linguistics}, pages 311--318, Philadelphia, Pennsylvania, USA, July 2002. Association for Computational Linguistics.
\newblock \doi{10.3115/1073083.1073135}.
\newblock \url{https://aclanthology.org/P02-1040/}.

\bibitem[Ryoo et~al.(2021)Ryoo, Piergiovanni, Arnab, Dehghani, and Angelova]{ryoo2021tokenlearner}
Michael~S. Ryoo, AJ~Piergiovanni, Anurag Arnab, Mostafa Dehghani, and Anelia Angelova.
\newblock Tokenlearner: adaptive space-time tokenization for videos.
\newblock In \emph{Proceedings of the 35th International Conference on Neural Information Processing Systems}, NIPS '21, Red Hook, NY, USA, 2021. Curran Associates Inc.
\newblock ISBN 9781713845393.

\bibitem[Vepa et~al.(2025)Vepa, Yu, Gan, Cuturrufo, Li, Wang, Scalzo, and Sun]{vepa2025brainmri}
Arvind~Murari Vepa, Yannan Yu, Jingru Gan, Anthony Cuturrufo, Weikai Li, Wei Wang, Fabien Scalzo, and Yizhou Sun.
\newblock A multimodal llm approach for visual question answering on multiparametric 3d brain mri.
\newblock \emph{arXiv preprint arXiv:2509.25889}, 2025.

\bibitem[Wen et~al.(2025)Wen, Gao, Li, He, and Zhang]{wen2025tokenpruning}
Zichen Wen, Yifeng Gao, Weijia Li, Conghui He, and Linfeng Zhang.
\newblock Token pruning in multimodal large language models: Are we solving the right problem?
\newblock \emph{arXiv preprint arXiv:2502.11501}, 2025.

\bibitem[Wu et~al.(2025)Wu, Zhang, Zhang, Hui, Wang, and Xie]{wu2025towards}
Chaoyi Wu, Xiaoman Zhang, Ya~Zhang, Hui Hui, Yanfeng Wang, and Weidi Xie.
\newblock Towards generalist foundation model for radiology by leveraging web-scale 2d\&3d medical data.
\newblock \emph{Nature Communications}, 16\penalty0 (1):\penalty0 7866, 2025.

\bibitem[Xu et~al.(2025)Xu, Chan, Li, Aljunied, Yuan, Wang, Xiao, Chen, Liu, Li, et~al.]{xu2025lingshu}
Weiwen Xu, Hou~Pong Chan, Long Li, Mahani Aljunied, Ruifeng Yuan, Jianyu Wang, Chenghao Xiao, Guizhen Chen, Chaoqun Liu, Zhaodonghui Li, et~al.
\newblock Lingshu: A generalist foundation model for unified multimodal medical understanding and reasoning.
\newblock \emph{arXiv preprint arXiv:2506.07044}, 2025.

\bibitem[Yang et~al.(2025{\natexlab{a}})Yang, Li, Yang, Zhang, Hui, Zheng, Yu, Gao, Huang, Lv, et~al.]{yang2025qwen3}
An~Yang, Anfeng Li, Baosong Yang, Beichen Zhang, Binyuan Hui, Bo~Zheng, Bowen Yu, Chang Gao, Chengen Huang, Chenxu Lv, et~al.
\newblock Qwen3 technical report.
\newblock \emph{arXiv preprint arXiv:2505.09388}, 2025{\natexlab{a}}.

\bibitem[Yang et~al.(2025{\natexlab{b}})Yang, Chen, Tian, Wang, Li, Yu, and Jia]{yang2025visionzip}
Senqiao Yang, Yukang Chen, Zhuotao Tian, Chengyao Wang, Jingyao Li, Bei Yu, and Jiaya Jia.
\newblock Visionzip: Longer is better but not necessary in vision language models.
\newblock In \emph{Proceedings of the IEEE/CVF Conference on Computer Vision and Pattern Recognition}, pages 19792--19802, 2025{\natexlab{b}}.

\bibitem[Zhang et~al.(2019)Zhang, Kishore, Wu, Weinberger, and Artzi]{zhang2019BERTScore}
Tianyi Zhang, Varsha Kishore, Felix Wu, Kilian~Q Weinberger, and Yoav Artzi.
\newblock Bertscore: Evaluating text generation with bert.
\newblock \emph{arXiv preprint arXiv:1904.09675}, 2019.

\end{thebibliography}
\clearpage
\beginappendix
\section{BreMRIs-VQA Dataset Details}
\label{app:bremris_dataset_statistics}

This section provides details for the construction, split protocol, and coverage of \dataset. Split-level statistics follow the released train/test partitions, while annotation-level distributions are computed from the raw generated QA pool.

\subsection{Dataset Source and Case-Level Split}

\dataset is built from multi-sequence breast MRI studies paired with clinical-expert-verified radiology and pathology reports. We split \dataset at the case level to prevent question--answer pairs from the same clinical case from appearing in both training and evaluation. 
For each case, we first build a distributional signature over its generated QA pairs, where each QA pattern is represented by the tuple \texttt{(task, question\_key, answer)} and counted across all questions belonging to that case. 
Because the full signature is high-dimensional, we compress it into a stable stratification label using the five most frequent QA patterns for that case. 
Cases with the same compressed label are grouped into the same stratum, shuffled with a fixed random seed of 42, and split using a 96\%/4\% train/test ratio. 
For strata containing only one case, where within-stratum splitting is impossible, cases are assigned after shuffling so that the global split ratio remains close to 96\%/4\%. 
The resulting case sets are disjoint by construction. 
The held-out case set is used as the test set for both open-ended and multiple-choice evaluation, so the two evaluation formats are matched at the clinical-case level and differ only in question format.

\subsection{Three-Stage Construction Pipeline}
\label{app:dataset_construction_detail}

\paragraph{Stage I: Structured clinical information extraction.}
The first construction stage converts free-text radiology and pathology reports into report-grounded structured representations. For radiology reports, the schema includes \texttt{organ\_level} attributes such as fibroglandular tissue (FGT), background parenchymal enhancement (BPE), symmetry, post-surgical changes, associated features, lymph nodes, BI-RADS category, and management recommendation. It also includes \texttt{lesion\_level} attributes for mass, non-mass enhancement, and non-enhancing lesions, covering side, quadrant, depth, size, morphology, and multi-sequence signal characteristics from T1w, T2w, DWI, ADC, and DCE. For pathology reports, the schema captures tumor location, histological type and grade, surgical margin status, lymph node involvement, molecular biomarkers, and pathological staging. All unsupported fields are set to \texttt{null}, schema-invalid outputs are discarded, and values outside the predefined clinical vocabulary are removed before question generation.

\paragraph{Stage II: Clinical question generation and task organization.}
Questions are generated from deterministic templates tied to structured clinical keys and organized into seven workflow-grounded tasks. \textit{Global background assessment} establishes diagnostic context through breast composition, BPE, symmetry, and post-surgical changes. \textit{Lesion detection and spatial localization} evaluates whether lesions are present and where they are located. \textit{Morphological characterization} covers descriptors of masses, non-mass enhancement, and non-enhancing lesions that support BI-RADS interpretation. \textit{Multi-modality functional reasoning} evaluates DWI/ADC signal, T2 signal, and DCE kinetic patterns that reflect diffusion restriction and enhancement behavior. \textit{Local invasion and nodal assessment} requires relational reasoning about surrounding structures and lymph nodes. \textit{Holistic diagnostic decision} integrates multiple imaging findings for BI-RADS categorization and management recommendation. \textit{Pathology prediction} bridges imaging phenotype with pathology-derived outcomes, including tumor characteristics, nodal status, biomarkers, and pathological staging. The complete key--task mapping is reported in Appendix~\ref{app:stage2_task}.

\paragraph{Stage III: Linguistic diversification.}
After template-based generation fixes the source key and answer label, Qwen3-235B~\cite{yang2025qwen3} proposes paraphrased questions to reduce surface-form bias. The LLM may change wording but cannot change the clinical key, answer label, answer type, or allowed answer space. For multiple-choice questions, the option label must remain unchanged, and paraphrases that leak the answer text or its synonyms are discarded. For open-ended questions, the paraphrased question is retained only if it asks for the same structured attribute as the original template. This stage increases linguistic coverage while preserving deterministic supervision from report-derived fields.

\subsection{Dataset Statistics and Task Taxonomy}

\begin{figure*}[t]
\centering
\includegraphics[width=\textwidth]{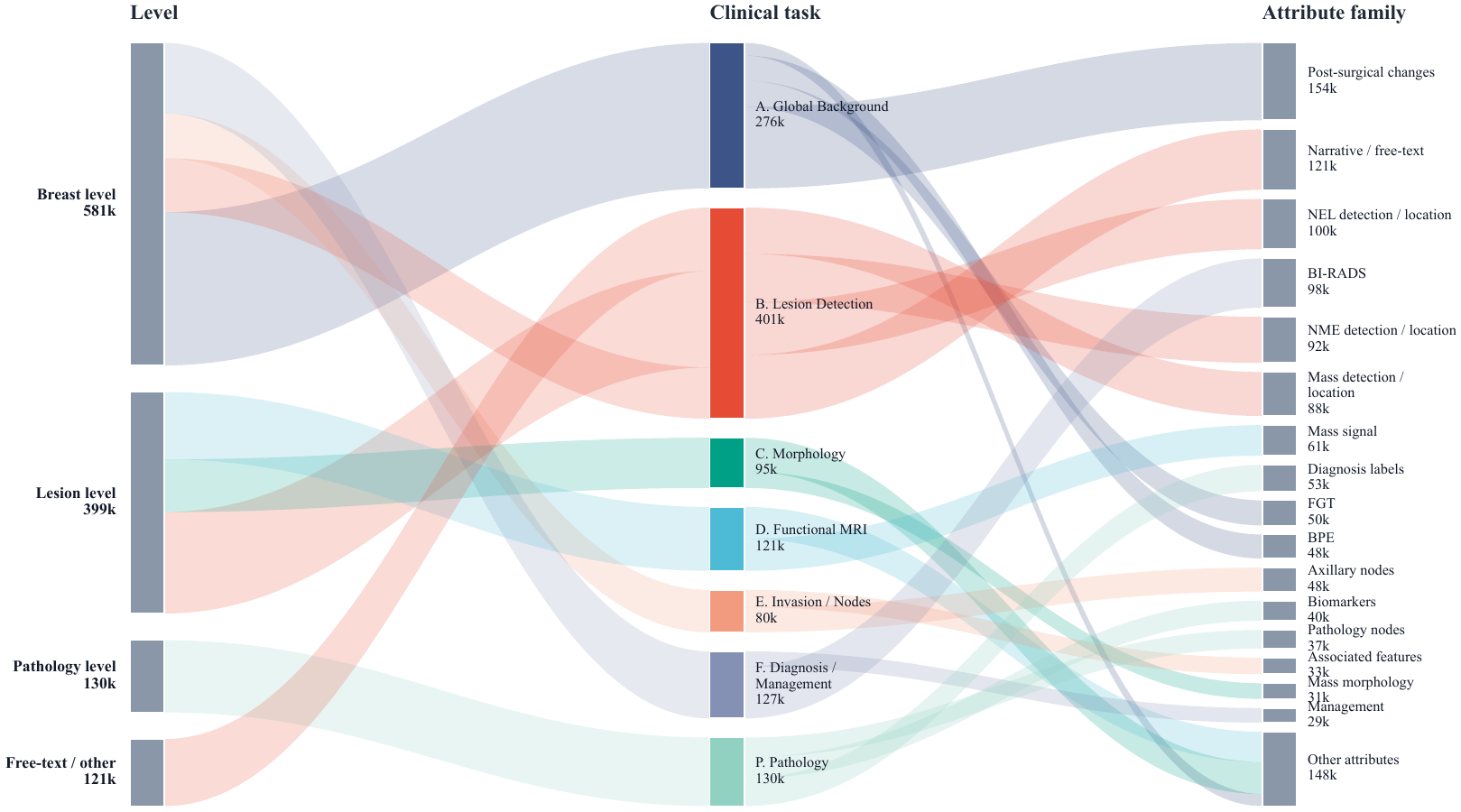}
\caption{Level--task--attribute flow of \dataset. Ribbon width is proportional to the number of generated QA templates, and colors correspond to the seven clinical workflow task groups.}
\label{fig:bremris_level_task_attribute_flow}
\end{figure*}

Figure~\ref{fig:bremris_level_task_attribute_flow} illustrates how generated QA templates are distributed across imaging levels, clinical task groups, and individual structured attributes.
Breast-level annotations yield the largest question volume because every case provides background tissue information spanning FGT composition, BPE level, breast symmetry, post-surgical changes, lymph node status, BI-RADS assessment, and management recommendation, producing dense per-case question sets regardless of lesion presence.
Lesion-level and pathology-level questions have narrower but complementary coverage, since they are generated only when corresponding structured findings are annotated for a given case.
Within each task group, the ribbon widths confirm that the label vocabulary is spread across multiple clinically meaningful attributes with no single attribute dominating, supporting the breadth and difficulty of evaluation that \dataset provides.

\begin{figure*}[t]
\centering
\includegraphics[width=\textwidth]{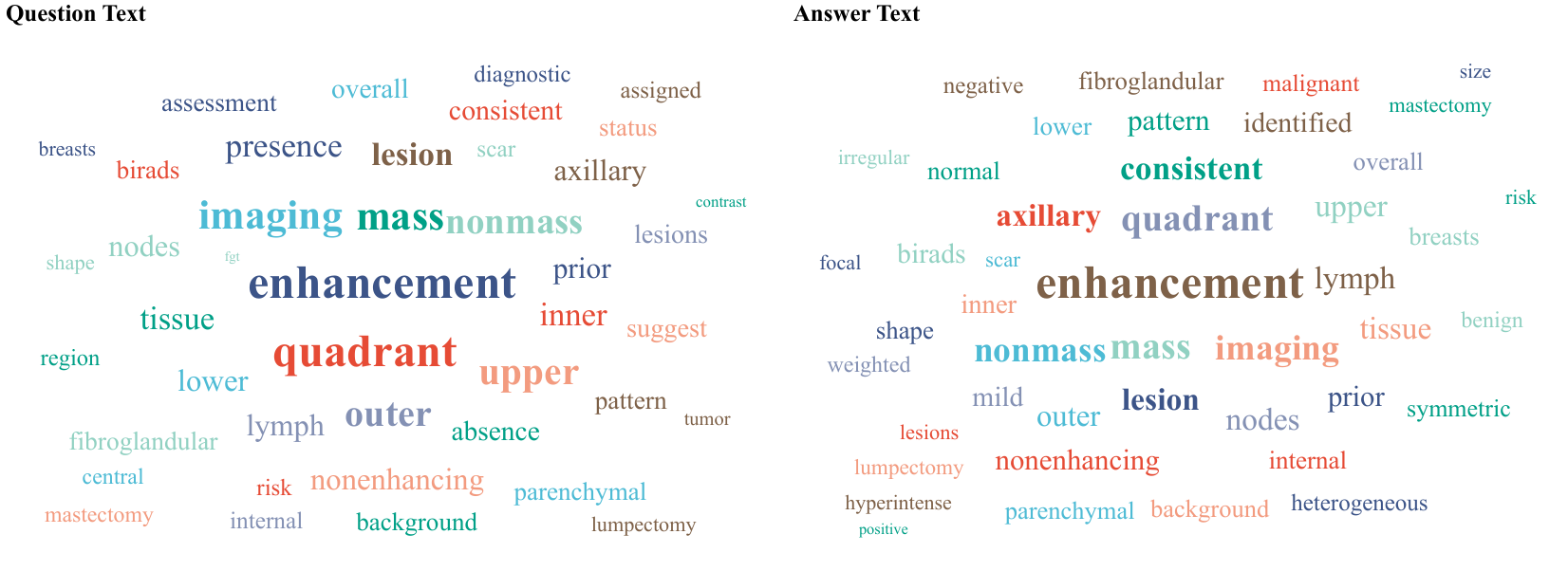}
\caption{Question and answer text word clouds for \dataset. Font size indicates token frequency after removing common function words. Unlike the attribute-frequency statistics, this visualization is computed directly from the natural-language question and answer text.}
\label{fig:bremris_question_answer_wordcloud}
\end{figure*}

Figure~\ref{fig:bremris_question_answer_wordcloud} shows word clouds computed from the natural-language question and answer text after the linguistic diversification step.
On the question side, the most frequent tokens are anatomical qualifiers and imaging descriptors (\textit{e.g.}, \textit{left}, \textit{right}, \textit{breast}, \textit{enhancement}), reflecting the structured attribute keys from which questions are derived.
On the answer side, the vocabulary is concentrated around standardized clinical categories (\textit{e.g.}, \textit{yes}, \textit{no}, \textit{minimal}, \textit{moderate}, \textit{positive}), consistent with the fixed answer spaces enforced during template-based generation.
The co-occurrence of rich question diversity with constrained, clinically grounded answers confirms that \dataset exercises linguistic generalization while maintaining deterministic supervision.
The table below details the per-task QA count and label distribution across all seven clinical workflow tasks.

\section{QA Generation and Quality Control}
\label{app:stage2_task}
\subsection{Template-Based Question Generation}

To construct clinically meaningful VQA tasks, we generate questions from a physician-designed schema covering breast-level context, lesion-level characterization, pathology findings, and free-text diagnostic reasoning. Each template is bound to a predefined source key and, when applicable, a fixed answer space, which keeps the generated QA pairs reproducible and grounded in report-derived evidence. The representative templates are grouped into breast-level, lesion-level, pathology-level, and free-text diagnostic questions below.

\begin{table*}[!htbp]
\centering
\renewcommand{\arraystretch}{0.94}
\resizebox{\linewidth}{!}{
\begin{tabular}{l|l|l}
\hline
\textbf{Category} &
  \multicolumn{1}{c|}{\textbf{Question Template}} &
  \multicolumn{1}{c}{\textbf{Options}} \\ \hline
\multirow{2}{*}{\begin{tabular}[c]{@{}l@{}}Fibroglandular\\ Tissue (FGT)\end{tabular}} &
  \begin{tabular}[c]{@{}l@{}}What is the fibroglandular tissue (FGT) category\\ of the left breast?\end{tabular} &
  A, B, C, D, None \\ \cline{2-3} 
 &
  \begin{tabular}[c]{@{}l@{}}What is the fibroglandular tissue (FGT) category \\ of the right breast?\end{tabular} &
  A, B, C, D, None \\ \hline
\multirow{2}{*}{\begin{tabular}[c]{@{}l@{}}Background\\ Parenchymal\\ Enhancement\end{tabular}} &
  \begin{tabular}[c]{@{}l@{}}What is the background parenchymal enhancement \\ (BPE) level of the left breast?\end{tabular} &
  \begin{tabular}[c]{@{}l@{}}Minimal, Mild, \\ Moderate, Marked\end{tabular} \\ \cline{2-3} 
 &
  \begin{tabular}[c]{@{}l@{}}What is the background parenchymal enhancement \\ (BPE) level of the right breast?\end{tabular} &
  \begin{tabular}[c]{@{}l@{}}Minimal, Mild, \\ Moderate, Marked\end{tabular} \\ \hline
Breast Symmetry &
  Are the breasts symmetric in size and shape? &
  Yes, No \\ \hline
\multirow{6}{*}{\begin{tabular}[c]{@{}l@{}}Post-surgical\\ Changes\end{tabular}} &
  Are there evidence of scar tissue in the left breast? &
  Yes, No \\ \cline{2-3} 
 &
  Are there evidence of scar tissue in the right breast? &
  Yes, No \\ \cline{2-3} 
 &
  \begin{tabular}[c]{@{}l@{}}Are there imaging findings consistent with prior\\  lumpectomy in the left breast?\end{tabular} &
  Yes, No \\ \cline{2-3} 
 &
  \begin{tabular}[c]{@{}l@{}}Are there imaging findings consistent with prior \\ lumpectomy in the right breast?\end{tabular} &
  Yes, No \\ \cline{2-3} 
 &
  \begin{tabular}[c]{@{}l@{}}Are there imaging findings consistent with prior \\ mastectomy in the left breast?\end{tabular} &
  Yes, No \\ \cline{2-3} 
 &
  \begin{tabular}[c]{@{}l@{}}Are there imaging findings consistent with prior \\ mastectomy in the right breast?\end{tabular} &
  Yes, No \\ \hline
\multirow{6}{*}{\begin{tabular}[c]{@{}l@{}}Associated\\ Imaging\\ Features\end{tabular}} &
  Is nipple retraction present in the left breast? &
  Yes, No \\ \cline{2-3} 
 &
  Is nipple retraction present in the right breast? &
  Yes, No \\ \cline{2-3} 
 &
  Is skin thickening present in the left breast? &
  Yes, No \\ \cline{2-3} 
 &
  Is skin thickening present in the right breast? &
  Yes, No \\ \cline{2-3} 
 &
  Is architectural distortion present in the left breast? &
  Yes, No \\ \cline{2-3} 
 &
  Is architectural distortion present in the right breast? &
  Yes, No \\ \hline
\multirow{4}{*}{\begin{tabular}[c]{@{}l@{}}Lymph Node\\ Assessment\end{tabular}} &
  Are the axillary left lymph nodes abnormal? &
  Yes, No \\ \cline{2-3} 
 &
  Are the axillary right lymph nodes abnormal? &
  Yes, No \\ \cline{2-3} 
 &
  Are the internal mammary left lymph nodes abnormal? &
  Yes, No \\ \cline{2-3} 
 &
  Are the internal mammary right lymph nodes abnormal? &
  Yes, No \\ \hline
\multirow{2}{*}{\begin{tabular}[c]{@{}l@{}}BI-RADS \\ Assessment\end{tabular}} &
  What is the BI-RADS assessment for the left breast? &
  0,1,2,3,4A,4B,4C,5,6 \\ \cline{2-3} 
 &
  What is the BI-RADS assessment for the right breast? &
  0,1,2,3,4A,4B,4C,5,6 \\ \hline
\multirow{2}{*}{\begin{tabular}[c]{@{}l@{}}Clinical \\ Management\end{tabular}} &
  Is follow-up recommended for this patient? &
  Yes, No \\ \cline{2-3} 
 &
  Is biopsy recommended for this patient? &
  Yes, No \\ \hline
\end{tabular}
}
\caption{Breast-level question templates used for MRI VQA generation.}\label{templates}
\label{tab:breast}
\end{table*}

\begin{table*}[!htbp]
\centering
\resizebox{\textwidth}{!}{
\begin{tabular}{l|l|l}
\toprule
\textbf{Category} &
  \multicolumn{1}{c}{\textbf{Question Template}} &
  \multicolumn{1}{c}{\textbf{Options}} \\ \hline
\multirow{4}{*}{\begin{tabular}[c]{@{}l@{}}Mass Presence \\ \& Location\end{tabular}} &
  \begin{tabular}[c]{@{}l@{}}Is there a mass in the upper inner quadrant \\ of the left breast?\end{tabular} &
  Yes, No \\ \cline{2-3} 
 &
  \begin{tabular}[c]{@{}l@{}}Is there a mass in the upper outer quadrant \\ of the left breast?\end{tabular} &
  Yes, No \\ \cline{2-3} 
 &
  \begin{tabular}[c]{@{}l@{}}Is there a mass in the lower inner quadrant \\ of the right breast?\end{tabular} &
  Yes, No \\ \cline{2-3} 
 &
  Is there a mass in the central region of the breast? &
  Yes, No \\ \hline
\multirow{2}{*}{\begin{tabular}[c]{@{}l@{}}Mass \\ Morphology\end{tabular}} &
  What is the shape of the mass in the left breast? &
  \begin{tabular}[c]{@{}l@{}}round, oval, \\ lobular, irregular\end{tabular} \\ \cline{2-3} 
 &
  What is the margin of the mass in the left breast? &
  \begin{tabular}[c]{@{}l@{}}circumscribed, irregular, \\ spiculated\end{tabular} \\ \hline
\multirow{3}{*}{\begin{tabular}[c]{@{}l@{}}Mass Signal \\ Characteristics\end{tabular}} &
  What is the T1-weighted signal of the mass? &
  \begin{tabular}[c]{@{}l@{}}hyperintense, isointense, \\ hypointense\end{tabular} \\ \cline{2-3} 
 &
  What is the T2-weighted signal of the mass? &
  \begin{tabular}[c]{@{}l@{}}hyperintense, isointense, \\ hypointense\end{tabular} \\ \cline{2-3} 
 &
  Is there diffusion restriction in the mass? &
  Yes, No \\ \hline
\multirow{2}{*}{\begin{tabular}[c]{@{}l@{}}Dynamic Enhan-\\ cement Kinetics\end{tabular}} &
  What is the initial enhancement kinetics of the mass? &
  Slow, Medium, Fast \\ \cline{2-3} 
 &
  \begin{tabular}[c]{@{}l@{}}What is the delayed phase enhancement \\ kinetics of the mass?\end{tabular} &
  \begin{tabular}[c]{@{}l@{}}Persistent, Plateau, \\ Wash-out\end{tabular} \\ \hline
\multirow{3}{*}{\begin{tabular}[c]{@{}l@{}}Non-mass \\ Enhancement\end{tabular}} &
  Is there a non-mass enhancement in the left breast? &
  Yes, No \\ \cline{2-3} 
 &
  \begin{tabular}[c]{@{}l@{}}What is the distribution pattern of \\ the non-mass enhancement?\end{tabular} &
  \begin{tabular}[c]{@{}l@{}}Focal, Linear, \\ Regional, Segmental, \\ Multiple regions, Diffuse\end{tabular} \\ \cline{2-3} 
 &
  \begin{tabular}[c]{@{}l@{}}What is the internal enhancement pattern of \\ the non-mass enhancement?\end{tabular} &
  \begin{tabular}[c]{@{}l@{}}Homogeneous, Heterogeneous, \\ Clustered, Clustered ring\end{tabular} \\ \hline
\multirow{2}{*}{\begin{tabular}[c]{@{}l@{}}Non-enhancing \\ Lesions\end{tabular}} &
  Is there a non-enhancing lesion in the breast? &
  Yes, No \\ \cline{2-3} 
 &
  What is the type of the non-enhancing lesion? &
  \begin{tabular}[c]{@{}l@{}}cyst, non-enhancing mass, \\ architectural distortion,\\ ductal precontrast high \\ signal on T1W\end{tabular} \\ \bottomrule
\end{tabular}
}
\caption{Lesion-level question templates used for MRI VQA generation.}
\label{tab:lesion}
\end{table*}

\begin{table*}[!htbp]
\centering
\resizebox{\textwidth}{!}{
\begin{tabular}{l|l|l}
\hline
\textbf{Category} &
  \multicolumn{1}{c|}{\textbf{Question Template}} &
  \multicolumn{1}{c}{\textbf{Options}} \\ \hline
\multirow{4}{*}{\begin{tabular}[c]{@{}l@{}}Presence \&\\ Surgical\\ Procedure\end{tabular}} &
  \begin{tabular}[c]{@{}l@{}}Is there a pathological lesion identified \\ in the left breast?\end{tabular} &
  Yes, No \\ \cline{2-3} 
 &
  \begin{tabular}[c]{@{}l@{}}Is there a pathological lesion identified \\ in the right breast?\end{tabular} &
  Yes, No \\ \cline{2-3} 
 &
  \begin{tabular}[c]{@{}l@{}}What surgical procedure was performed \\ on the left breast?\end{tabular} &
  \begin{tabular}[c]{@{}l@{}}Lumpectomy, \\ Mastectomy, Other\end{tabular} \\ \cline{2-3} 
 &
  \begin{tabular}[c]{@{}l@{}}What surgical procedure was performed \\ on the right breast?\end{tabular} &
  \begin{tabular}[c]{@{}l@{}}Lumpectomy, \\ Mastectomy, Other\end{tabular} \\ \hline
\multirow{2}{*}{\begin{tabular}[c]{@{}l@{}}Tumor\\ Characteristics\end{tabular}} &
\begin{tabular}[c]{@{}l@{}}Is the maximum pathological tumor size\\  larger than 2 cm?\end{tabular} &
  Yes, No \\ \cline{2-3} 
 &
  What is the histological grade of the tumor? &
  1, 2, 3, Unknown \\ \hline
\multirow{2}{*}{\begin{tabular}[c]{@{}l@{}}Surgical Margins\\ \& Lymph Nodes\end{tabular}} &
  Are the surgical margins involved by tumor? &
  Positive, Negative \\ \cline{2-3} 
 &
  Are lymph nodes involved based on pathology? &
  Positive, Negative \\ \hline
\multirow{6}{*}{\begin{tabular}[c]{@{}l@{}}Biomarkers \&\\ Pathological\\ Staging\end{tabular}} &
  What is the ER status of the tumor? &
  \begin{tabular}[c]{@{}l@{}}Positive, Negative, \\ Equivocal, Unknown\end{tabular} \\ \cline{2-3} 
 &
  What is the PR status of the tumor? &
  \begin{tabular}[c]{@{}l@{}}Positive, Negative, \\ Equivocal, Unknown\end{tabular} \\ \cline{2-3} 
 &
  What is the HER2 status of the tumor? &
  \begin{tabular}[c]{@{}l@{}}Positive, Negative, \\ Equivocal, Unknown\end{tabular} \\ \cline{2-3} 
 &
  What is the Ki67 status of the tumor? &
  \begin{tabular}[c]{@{}l@{}}Positive, Negative, \\ Equivocal, Unknown\end{tabular} \\ \cline{2-3} 
 &
  What is the AR status of the tumor? &
  \begin{tabular}[c]{@{}l@{}}Positive, Negative, \\ Equivocal, Unknown\end{tabular} \\ \cline{2-3} 
 &
  What is the pathological TNM stage? &
  \begin{tabular}[c]{@{}l@{}}Stage 0, Stage I, Stage II, \\ Stage III, Stage IV, Unknown\end{tabular} \\ \hline
\end{tabular}
}
\caption{Pathology question templates used for pathology report VQA generation.}
\label{tab:pathology}
\end{table*}

\begin{table}[tbp]
\centering
\small
\begin{tabular}{@{}p{0.95\linewidth}@{}}
\hline
\multicolumn{1}{c}{\textbf{Question Template}} \\ \hline
Describe the overall fibroglandular tissue composition and background parenchymal enhancement of both breasts. \\ \hline
Provide a comprehensive description of the mass identified in the breast MRI. \\ \hline
Describe the imaging characteristics of the non-mass enhancement in the breast. \\ \hline
Describe the appearance of axillary and internal mammary lymph nodes. \\ \hline
Provide the final BI-RADS assessment and overall diagnostic impression. \\ \hline
\end{tabular}
\caption{Free-text diagnostic question templates used for holistic clinical reasoning.} 
\label{tab:free}
\end{table}

Tables~\ref{tab:breast}--\ref{tab:free} summarize representative templates. Breast-level questions cover global context such as FGT, BPE, symmetry, post-surgical changes, lymph nodes, BI-RADS assessment, and management. Lesion-level questions cover detection, location, morphology, multi-sequence signal, enhancement kinetics, non-mass enhancement, and non-enhancing lesions. Pathology-level questions cover surgical procedures, tumor characteristics, margins, lymph nodes, biomarkers, and pathological staging. Free-text questions simulate holistic diagnostic reporting and require models to integrate multiple findings into clinically meaningful descriptions.

\subsection{Clinical Workflow Task Organization}
Although the previous step produces individual VQA samples from structured annotations, clinical diagnosis is inherently a multi-stage reasoning process rather than a collection of isolated attribute predictions. Radiologists typically interpret breast MRI by progressively integrating information across different levels of analysis, starting from global contextual assessment and lesion detection, and eventually arriving at a holistic diagnostic conclusion. To better reflect this real diagnostic workflow, we organize the generated questions into a hierarchy of clinically meaningful tasks. Instead of grouping questions purely according to low-level visual attributes, we introduce a rule-based \textit{key--task mapping} mechanism designed in collaboration with clinicians. Specifically, each generated question is associated with a conceptual identifier called a \texttt{question\_key}, which corresponds to the structured attribute from which the question is derived. These keys provide a unified representation for linking structured annotations, generated questions, and downstream evaluation tasks. A predefined mapping table then assigns each \texttt{question\_key} to a specific diagnostic task using either exact matching or prefix-based rules. Exact rules match a single attribute key, whereas prefix rules group a family of semantically related attributes under the same task category.
Through this mechanism, all generated questions are systematically organized into a set of clinically motivated reasoning tasks that correspond to the major stages of breast MRI interpretation. The mapping rules used to assign questions to tasks are summarized in Table~\ref{tab:key_task_mapping}. Concretely, the task taxonomy consists of the following stages:

\paragraph{Global Background Assessment. } 
Radiologists first evaluate the overall background context of the breast before focusing on specific lesions. This stage includes questions related to fibroglandular tissue composition, background parenchymal enhancement, breast symmetry, and post-surgical changes. These factors influence lesion visibility and diagnostic difficulty.

\paragraph{Lesion Detection and Localization. } 
Once the global context is established, the next step is to determine whether suspicious findings are present and where they are located. Questions in this category involve detecting masses, non-mass enhancement (NME), and non-enhancing lesions, as well as identifying their anatomical locations within the breast.

\paragraph{Morphological Characterization. }
After lesion detection, radiologists analyze morphological features to assess malignancy risk. This stage includes attributes such as lesion size, shape, margins, and distribution patterns, which correspond to key descriptors used in standardized breast imaging reporting systems.

\paragraph{Multi-Modal Functional Reasoning. }
Breast MRI provides multiple imaging modalities that reveal complementary physiological information. This task group evaluates functional characteristics such as signal intensity across sequences and contrast enhancement patterns, requiring models to integrate multi-modal cues.

\paragraph{Local Invasion and Nodal Reasoning. }
In addition to lesion features, clinicians also examine signs of local tissue invasion and lymph node involvement in diagnosis. Questions in this category involve associated features such as nipple retraction, skin thickening, architectural distortion and lymph node abnormalities.

\paragraph{Holistic Diagnosis. }
The final stage of radiological reasoning involves synthesizing all available information to produce a diagnostic assessment and clinical recommendation. This task includes predicting BI-RADS categories and suggested clinical management decisions.

\paragraph{Pathology Prediction. }
Beyond imaging-based reasoning, we further introduce pathology-oriented tasks derived from structured pathology reports. These questions evaluate the model's ability to infer pathological outcomes, including tumor characteristics, surgical margins, lymph node status, biomarker expression, and pathological staging.
\begin{table}[t]
\centering
\setlength{\tabcolsep}{2pt}
\renewcommand{\arraystretch}{0.92}
\begin{tabular}{lcccc}
\toprule
Task & Fact. & Clin. & Overall & Pass (\%) \\
\midrule
Global background & 2.94 & 2.92 & 2.93 & 99.2 \\
Lesion localization & 2.91 & 2.89 & 2.90 & 98.6 \\
Morphology & 2.86 & 2.83 & 2.84 & 97.5 \\
Functional reasoning & 2.79 & 2.75 & 2.77 & 96.2 \\
Local invasion/nodal & 2.88 & 2.86 & 2.87 & 97.9 \\
Holistic diagnosis & 2.82 & 2.79 & 2.80 & 96.8 \\
Pathology prediction & 2.80 & 2.82 & 2.81 & 96.5 \\
\bottomrule
\end{tabular}
\caption{Quality-control summary for generated QA pairs. Fact., Clin., and Overall denote factual consistency, clinical validity, and overall report support on a 0--3 scale. Pass rate denotes the percentage of expert-audited QA pairs that pass answer--evidence checking.}
\label{tab:quality_control}
\end{table}
\begin{table*}[t]
\centering
\resizebox{\textwidth}{!}{\begin{tabular}{c|l|l}
\toprule
\textbf{Task Name} &
  \textbf{Exact Keys} &
  \textbf{Prefixes} \\ \midrule
\begin{tabular}[c]{@{}c@{}}Global Background\\ Assessment\end{tabular} &
  breast\_level.breast\_symmetry &
  \begin{tabular}[c]{@{}l@{}}breast\_level.FGT., breast\_level.BPE., \\ breast\_level.post\_surgical\_changes.\end{tabular} \\ \hline
\begin{tabular}[c]{@{}c@{}}Lesion Detection \\ \& Localization\end{tabular} &
  \begin{tabular}[c]{@{}l@{}}mass, non\_mass\_enhancement,\\ non\_enhancing\_lesion, mass.count,\\ non\_mass\_enhancement.count, \\ non\_enhancing\_lesion.count\end{tabular} &
  \begin{tabular}[c]{@{}l@{}}mass.location., \\ non\_mass\_enhancement.location.,   \\ non\_enhancing\_lesion.location.\end{tabular} \\ \hline
\begin{tabular}[c]{@{}c@{}}Morphological \\ Characterization\end{tabular} &
  \begin{tabular}[c]{@{}l@{}}non\_mass\_enhancement.distribution,   \\ non\_enhancing\_lesion.lesion\_type\_detail\end{tabular} &
  \begin{tabular}[c]{@{}l@{}}mass.size\_mm., mass.morphology., \\ non\_mass\_enhancement.size\_mm.,   \\ non\_enhancing\_lesion.size\_mm.\end{tabular} \\ \hline
\begin{tabular}[c]{@{}c@{}}Multi-Modal \\ Functional Reasoning\end{tabular} &
  \begin{tabular}[c]{@{}l@{}}non\_mass\_enhancement.\\ internal\_enhancement\_pattern\end{tabular} &
  \begin{tabular}[c]{@{}l@{}}mass.signal\_characteristics.,  \\ non\_mass\_enhancement.signal\_characteristics.,   \\ non\_enhancing\_lesion.signal\_characteristics., \\ mass.internal\_enhancement.\end{tabular} \\ \hline
\begin{tabular}[c]{@{}c@{}}Local Invasion \&\\  Nodal Reasoning\end{tabular} &
  - &
  \begin{tabular}[c]{@{}l@{}}breast\_level.associated\_features.,   \\ breast\_level.lymph\_nodes.\end{tabular} \\ \hline
Holistic Diagnosis &
  - &
  \begin{tabular}[c]{@{}l@{}}breast\_level.BI-RADS., \\ breast\_level.management.\end{tabular} \\ \hline
Pathology Prediction &
  - &
  pathology. \\ \bottomrule
\end{tabular}
}
\caption{Key--task mapping rules used to organize generated questions into clinical-workflow based tasks.}
\label{tab:key_task_mapping}
\end{table*}

\subsection{LLM Verification and Human Expert Consistency Check}
\label{app:dataset_qc_detail}
Each generated QA candidate is verified against the source report evidence with a radiologist-style quality-control prompt. The verifier assigns three 0--3 scores with metric-specific rubrics. For factual consistency, 0 means the answer contradicts or is absent from the report, 1 means only weak or indirect support, 2 means partial support with missing qualifiers such as location, laterality, or pathology status, and 3 means direct report support without hallucinated findings. For clinical validity, 0 means clinically implausible or radiologically incorrect, 1 means medically ambiguous or poorly grounded, 2 means clinically plausible with minor imprecision, and 3 means radiologically coherent and clinically correct. For overall support, 0 means the QA pair should be discarded, 1 means it requires manual review, 2 means it is usable but partially supported, and 3 means it can be retained as fully report-grounded supervision. Candidates with weak support are removed or sent for manual review. The full verification prompt is provided in Appendix~\ref{app:prompt_veri}. 
In addition to automatic verification, two board-certified breast-imaging radiologists independently review a stratified sample of 3{,}500 QA pairs (500 per task) and mark whether each pair passes answer--evidence checking, where a pass means that the answer can be justified by the cited source evidence without requiring additional assumptions. Failed cases are used to refine templates and filtering rules.

\section{Background}
\textbf{Medical Vision-Language Models}.
Recent medical vision-language models extend general multi-modal LLMs to clinical images and reports, enabling stronger medical understanding and generation~\cite{blankemeier_kumar2026merlin,xu2025lingshu}. 
A clear trend is to scale training data and align multi-modal reasoning for broad clinical tasks~\cite{hu2024omnimedvqa}. For example, HealthGPT~\cite{lin2025healthgpt} and Lingshu~\cite{xu2025lingshu} emphasize unified training for medical comprehension and generation. 
In radiology, RadFM~\cite{wu2025towards} leverages large-scale 2D and 3D medical data and supports volumetric inputs, providing a strong generalist baseline for 3D medical understanding. 
Despite these advances, most medical MLLMs still adopt dense visual tokenization and expose large numbers of tokens to the decoder, which can be problematic for 3D volumes: volumetric scans contain heavy redundancy while clinically relevant cues are sparse. This motivates our focus on \emph{selective visual exposure} to reduce signal dilution.

\noindent
\textbf{3D Medical VQA and Benchmarks}.
3D medical VQA requires cross-slice spatial understanding and fine-grained evidence aggregation, making it harder than 2D VQA. 
Existing 3D medical MLLMs~\cite{hamamci2024ct2rep, lin2026omnict} and benchmarks~\cite{hamamci2026ct-rate, gai20253drad} have taken important steps toward this goal. M3D~\cite{bai2024m3d} advances 3D multimodal learning with large-scale 3D data and benchmarks that cover VQA and localization. 
Photon~\cite{fang2026photon} studies volume understanding with variable-length volumetric tokenization and training strategies that speed up volume understanding. Closer to multi-modal MRI, mpLLM~\cite{vepa2025brainmri} introduces a prompt-conditioned hierarchical MoE for VQA over multiparametric 3D brain MRI, using modality- and token-level experts to fuse interrelated MRI modalities.  
On the evaluation side, 3D-RAD emphasizes diverse 3D radiology VQA tasks with multi-temporal settings~\cite{gai20253drad}, while DeepTumorVQA targets tumor-centric 3D reasoning with expert-level questions across recognition, measurement, and clinical reasoning~\cite{chen2025deeptumor}. 
These efforts highlight a shared challenge: 3D medical image reasoning needs both global context and subtle local cues, yet dense token exposure can dilute the sparse diagnostic signals. Moreover, most public benchmarks focus on CT and single-volume inputs, systematic evaluation on \emph{multi-modal MRI} scans with workflow-grounded questions remains limited, motivating our \dataset benchmark and the \model framework.

\noindent
\textbf{Visual Token Reduction and Resampling}.
MLLMs often encode images into many visual tokens, motivating token reduction, token merging, and learned resampling. Perceiver-style resamplers in Flamingo~\cite{alayrac2022flamingo} compress visual features into a fixed number of latent tokens before language decoding, while TokenLearner~\cite{ryoo2021tokenlearner}, Token Merging~\cite{bolya2023token}, and FastV~\cite{Chen2024fastv} reduce visual computation through adaptive token selection or merging. More recent MLLM pruning studies further analyze whether reduced visual token budgets preserve reasoning quality~\cite{wen2025tokenpruning,alvar2025divprune,yang2025visionzip}. The closest design to our first-stage selector is MMTok~\cite{dong2026mmtok}, which formulates visual token selection as coverage maximization. We therefore do not claim the coverage objective itself as the main novelty. Instead, \model adapts coverage-based selection as a modality-wise redundancy reducer for 3D medical volumes and couples it with language-conditioned gated retrieval and marginal-utility regularization, which are designed for multi-sequence volumetric VQA rather than generic 2D VLM acceleration.

\section{Prompt Templates for VQA Data Generation}
\label{app:stage1_prompt}
\definecolor{PromptTeal}{HTML}{2A7F7F}
\definecolor{PromptAmber}{HTML}{9B6500}
\definecolor{PromptWine}{HTML}{9A1B4F}
\definecolor{PromptRed}{HTML}{B7322C}
\definecolor{PromptPurple}{HTML}{6E2A7E}
\definecolor{PromptBlue}{HTML}{2D7F9F}

\newtcolorbox{promptbox}[3][]{enhanced,
  breakable,
  colback=#3!5,
  colframe=#3!72!black,
  colbacktitle=#3!75!black,
  coltitle=white,
  fonttitle=\bfseries,
  title={#2},
  attach boxed title to top left={xshift=2mm,yshift=-2mm},
  boxed title style={
    colback=#3!75!black,
    colframe=#3!75!black,
    sharp corners,
    boxrule=0pt,
    left=2mm,
    right=2mm,
    top=0.6mm,
    bottom=0.6mm
  },
  arc=1mm,
  boxrule=0.7pt,
  top=1.5mm,
  left=2mm,
  right=2mm,
  bottom=1mm,
  before skip=0.15em,
  after skip=0.15em,
  #1
}
\setlength{\parskip}{2pt}
\subsection{Report-to-JSON Extraction Prompt Text}
\begin{promptbox}{Breast MRI Report-to-JSON Extraction}{PromptTeal}
\begin{Verbatim}[fontsize=\footnotesize, breaklines=true, breakanywhere=true]
You are a breast imaging expert with extensive clinical experience.
Your task: extract all structured information from a free-text breast MRI report into a standardized JSON.
All values must be strictly grounded in the input report. Do NOT add any inference or medical common sense.
If a field is not explicitly mentioned, set it to null (do not guess).

---------------------------
Core Requirements
---------------------------
1) Output must be a valid JSON only. No explanations.
2) The JSON schema must exactly match the provided template. Do not rename, remove, or add fields.
3) All values must be extracted from the report text. If not mentioned, use null.
   - Example: if "skin thickening" is not stated, set it to null (not false).
   - If enhancement kinetics is not described, set it to null.
4) Convert "cm" to millimeters (mm).
5) If multiple lesions are described, create multiple entries and place them in the correct arrays:
   - mass, non_mass_enhancement, non_enhancing_lesion
6) If a lesion type is not present, output an empty array [] for that type.

---------------------------
FGT (Fibroglandular Tissue) Rules
---------------------------
Map keywords to FGT grade:
- "A", "almost entirely fatty"  -> "A"
- "B", "scattered fibroglandular" -> "B"
- "C", "heterogeneously dense" -> "C"
- "D", "extremely dense" -> "D"
If left/right differ, fill both sides separately.
If not described, set FGT.left = null and FGT.right = null.

---------------------------
Lesion Type Rules (Strict)
---------------------------
Mass:
- explicit bounded lesion such as "mass", "nodule", etc.
Non-mass enhancement (NME):
- regional/linear/segmental/ductal enhancement without a clear 3D boundary.
Non-enhancing lesion:
- lesions described as non-enhancing (e.g., cystic signal, pre-contrast high signal on T1, architectural distortion without enhancement).
Classify strictly based on report semantics.

---------------------------
Kinetics Rules
---------------------------
Initial phase:
- Fast: early marked enhancement / rapid peak
- Medium: moderate enhancement
- Slow: mild or minimal early enhancement
Delayed phase:
- Persistent / Plateau / Wash-out according to report wording
If kinetics not described, set to null.

---------------------------
Location Parsing
---------------------------
Quadrant:
- UIQ / UOQ / LIQ / LOQ / central
Depth:
- anterior / middle / posterior
If unclear, set to null.

---------------------------
BI-RADS Rules
---------------------------
If BI-RADS is explicitly provided for left/right, fill accordingly.
If only a single BI-RADS is given without side, fill both.
If not described, set to null.

---------------------------
Final Output
---------------------------
Read the breast MRI report text provided next.
Extract all information and output JSON only.
\end{Verbatim}
\end{promptbox}
\subsection{Breast MRI Report JSON Schema}
\begin{promptbox}{Report JSON Schema}{PromptBlue}
\begin{Verbatim}[fontsize=\scriptsize, breaklines=true, breakanywhere=true]
{
  "breast_level": {
    "FGT": {
      "left": "C",  // or "A", "B", "C", "D"
      "right": "C"  // or "A", "B", "C", "D"
    },
    "BPE": {
      "left": "Moderate",  // or "Minimal", "Mild", "Moderate", "Marked"
      "right": "Mild"
    },
    "breast_symmetry": true,
    "post_surgical_changes": {
      "left": {
        "scar_tissue": false,
        "lumpectomy_changes": false,
        "mastectomy_changes": false
      },
      "right": {
        "scar_tissue": false,
        "lumpectomy_changes": false,
        "mastectomy_changes": false
      }
    },
    "associated_features": {
      "left": {
        "nipple_retraction": null,
        "nipple_invasion": null,
        "skin_retraction": null,
        "skin_thickening": null,
        "skin_invasion": {
          "direct_invasion": null,
          "inflammatory_cancer": null
        },
        "axillary_adenopathy": null,
        "pectoralis_muscle_invasion": null,
        "chest_wall_invasion": null,
        "architectural_distortion": null,
        "ductal_dilation": null  // or "mild", "marked", null
      },
      "right": {
        "nipple_retraction": null,
        "nipple_invasion": null,
        "skin_retraction": null,
        "skin_thickening": null,
        "skin_invasion": {
          "direct_invasion": null,
          "inflammatory_cancer": null
        },
        "axillary_adenopathy": null,
        "pectoralis_muscle_invasion": null,
        "chest_wall_invasion": null,
        "architectural_distortion": null,
        "ductal_dilation": null  // or "mild", "marked", null
      }
    },
    "lymph_nodes": {
      "axillary_left": null,  // or "normal", "abnormal"
      "axillary_right": null,
      "internal_mammary_left": null,
      "internal_mammary_right": null
    },
    "BI-RADS": {
      "left": null,  // or "0", "1", "2", "3", "4A", "4B", "4C", "5", "6"
      "right": null
    },
    "management": {
      "follow_up_recommended": null,  // or true, false
      "biopsy_recommended": null
    }
  },
  "mass": [
    {
      "id": 1,
      "side": "left",
      "location": {
        "quadrant": "UOQ",  // or "UIQ","UOQ","LIQ","LOQ","central", null
        "depth": "middle"  // or "anterior","middle","posterior", null
      },
      "size_mm": {
        "long": 12,
        "short": 9,
        "depth": 8
      },
      "signal_characteristics": {
        "T1W": "hyperintense",  // or "hyperintense","isointense","hypointense",null
        "T2W": "hyperintense",  // or "hyperintense","isointense","hypointense",null
        "DWI_restriction": "present",  // or "present","absent",null
        "ADC": "hyperintense",  // or "hyperintense","isointense","hypointense",null
        "kinetics": {
          "initial": "Fast",
          "delayed": "Wash-out"
        }
      },
      "morphology": {
        "shape": "oval",  // or "round","oval","lobular","irregular",null
        "margin": "circumscribed"  // or "circumscribed","irregular","spiculated",null
      },
      "internal_enhancement": {
        "pattern": "heterogeneous",  // or "homogeneous","heterogeneous","rim_enhancement","dark_internal_septations",null
        "associated_vascularity": {
          "adjacent_vessel_sign": null,
          "increased_peritumoral_vascularity": null
        }
      }
    }
  ],
  "non_mass_enhancement": [
    {
      "lesion_id": 1,
      "side": "left",
      "location": {
        "quadrant": null,
        "depth": null
      },
      "size_mm": {
        "long": null,
        "short": null,
        "depth": null
      },
      "distribution": "Segmental",  // or "Focal","Linear","Regional","Segmental","Multiple regions","Diffuse",null
      "internal_enhancement_pattern": "Clustered_ring",  // or "Homogeneous","Heterogeneous","Clustered","Clustered_ring",null
      "signal_characteristics": {
        "T1W": "hyperintense",  // or "hyperintense","isointense","hypointense",null
        "T2W": "hyperintense",  // or "hyperintense","isointense","hypointense",null
        "DWI_restriction": "present",  // or "present","absent",null
        "ADC": "hyperintense",  // or "hyperintense","isointense","hypointense",null
        "kinetics": {
          "initial": "Fast",  // "Slow","Medium","Fast",null
          "delayed": "Wash-out"  // "Persistent","Plateau","Wash-out",null
        }
      }
    }
  ],
  "non_enhancing_lesion": [
    {
      "lesion_id": 1,
      "side": "left",
      "location": {
        "quadrant": null,
        "depth": null
      },
      "size_mm": {
        "long": null,
        "short": null,
        "depth": null
      },
      "signal_characteristics": {
        "T1W": "hyperintense",  // or "hyperintense","isointense","hypointense",null
        "T2W": "hyperintense",  // or "hyperintense","isointense","hypointense",null
        "DWI_restriction": "present",  // or "present","absent",null
        "ADC": "hyperintense",  // or "hyperintense","isointense","hypointense",null
        "kinetics": {
          "initial": "Fast",  // "Slow","Medium","Fast",null
          "delayed": "Wash-out"  // "Persistent","Plateau","Wash-out",null
        }
      },
      "lesion_type_detail": "ductal_precontrast_high_signal_on_T1W"  // or "cyst","non_enhancing_mass","architectural_distortion",null
    }
  ]
}
\end{Verbatim}
\end{promptbox}
\subsection{Pathology Report Extraction Prompt Text}
\begin{promptbox}{Pathology Report-to-JSON Extraction}{PromptPurple}
\begin{Verbatim}[fontsize=\footnotesize, breaklines=true, breakanywhere=true]
# Role
You are an expert Pathologist and Data Structuring Specialist. Your task is to extract information from raw Chinese Breast Pathology Reports and convert it into a structured **English JSON** format.

# Input Data Description
1. Input data format:
   {
      'type': Surgical pathology | Non-surgical pathology,
      'pathology': Pathology Reports.
   }
2. The **Surgical pathology**: text fields separated by spaces or colons.
3. The **Non-surgical pathology**: often formatted as Python set strings (e.g., `"{'text', 'text'}"`).
4. **Symbols**: `(-)` means Negative, `(+)` means Positive, `/` often means Not Applicable or Not Found.

# Translation Rules (Strict Adherence)
Use the following mapping to translate Chinese terms to standard English medical terminology:

- Invasive carcinoma -> "Invasive Carcinoma"
- No special type (NST) -> "Invasive Carcinoma of No Special Type (NST)"
- Ductal carcinoma in situ -> "Ductal Carcinoma In Situ (DCIS)"
- Fibroadenoma -> "Fibroadenoma"
- Sclerosing adenosis -> "Sclerosing Adenosis"
- Mastopathy -> "Mastopathy" or "Fibrocystic Changes"
- Spindle cell lesion -> "Spindle Cell"

## Anatomy & Margins
- Left / Right -> "Left" / "Right"
- Medial / Lateral / Superior / Inferior -> "Medial" / "Lateral" / "Superior" / "Inferior"
- Basal / Superficial -> "Basal" / "Superficial"
- Quadrant (UIQ / UOQ / LIQ / LOQ) -> "Quadrant" (UIQ, UOQ, LIQ, LOQ)
- Margin -> "Margin"

## Molecular & IHC
- Immunohistochemistry -> "IHC"
- Strong / Moderate / Weak -> "Strong" / "Moderate" / "Weak"
- Negative / Positive -> "Negative" / "Positive"
- Amplification -> "Amplification"

# Data Parsing Rules
1. **Margins**: Extract the distance (cm) if available. Convert `(-)` to `Negative`.
2. **IHC**: Look for Supplementary Report or Immunohistochemistry sections. Merge them into the `biomarkers` field.
3. **Measurements**: Convert all dimensions to `cm`. Extract the largest dimension as `max_dimension_cm`.
4. For **Non-surgical pathology**: Clean the Python set syntax (`{'...'}`) and combine the text before extraction.

# Final Output
Return strictly a JSON object following the schema below. Do not add explanations.
\end{Verbatim}
\end{promptbox}
\subsection{Pathology Report JSON Schema}
\begin{promptbox}{Surgical Pathology JSON Schema}{PromptBlue}
\begin{Verbatim}[fontsize=\scriptsize, breaklines=true, breakanywhere=true]
{
  "left": {
    "is_present": boolean,
    "procedure": "string (e.g., BCS, Mastectomy, Biopsy)",
    "lesions": [
      {
        "id": 1,
        "location_detail": "string",
        "diagnosis": {
          "main_type": "string",
          "grade": "string (e.g., Grade II)",
          "full_text": "string (translated summary)"
        },
        "size": {
          "original_text": "string",
          "max_dim_cm": float
        },
        "margins": {
          "status": "Negative" | "Positive" | "Unknown",
          "closest_distance_cm": float,
          "details": "string (e.g., Superior margin 2.5cm)"
        },
        "biomarkers_ihc": {
           "ER": boolean, "PR": boolean, "HER2": boolean, "Ki67": boolean, "AR": boolean, "notes": boolean
        }
      }
    ],
    "lymph_nodes": {
      "status": "Positive" | "Negative" | "Not Performed",
      "positive_count": int,
      "total_count": int,
      "sentinel_positive": int, 
      "sentinel_total": int
    },
    "tnm_stage": "string"
  },
  "right": {
    // Exact same structure as "left", or null if not present
    "is_present": boolean,
    "procedure": "string",
    "lesions": [],
    "lymph_nodes": {},
    "tnm_stage": "string"
  }
}
\end{Verbatim}
\end{promptbox}
\subsection{QA Paraphrasing Prompt}
\begin{promptbox}{Open-VQA Generator Prompt}{PromptAmber}
\begin{Verbatim}[fontsize=\footnotesize, breaklines=true, breakanywhere=true]
You are a medical imaging data assistant specializing in breast MRI.
Your task is to rewrite a structured attribute into a natural-language Question-Answer pair suitable for a Breast MRI VQA dataset. 

# Rules
1. The Question MUST be open-ended and descriptive.
   - Do NOT ask yes/no questions.
   - Do NOT include options or binary choices.
2. The Answer MUST be a complete sentence that:
   - Is strictly based on the provided "Answer" value.
   - Does NOT introduce any new medical interpretation, diagnosis, or inference.
3. Do NOT introduce information not explicitly contained in the input.
4. Do NOT repeat or list the provided options in the output.
5. Use varied sentence structures across different samples.
6. Use precise anatomical and imaging terminology appropriate for breast MRI.
7. Output ONLY valid JSON in the specified format. No extra text.

# Input (JSON)
{
  "Question": "...",
  "options": ["...", "...", "..."],
  "Answer": "..."
}

# Output Format (JSON)
{
  "Question": "...",
  "Answer": "..."
}
\end{Verbatim}
\end{promptbox}

\subsection{LLM Verification Prompt}\label{app:prompt_veri}
\begin{promptbox}{Quality-Control Prompt}{PromptRed}
\begin{Verbatim}[fontsize=\footnotesize, breaklines=true, breakanywhere=true]
You are an expert breast MRI radiologist. Your task is to evaluate whether the provided Question-Answer pair is factually supported by the given breast MRI evidence. Evaluate the QA pair across the following dimensions:

1. Factual Consistency
Does the answer accurately reflect the MRI evidence without unsupported findings?

2. Clinical Validity
Is the answer clinically reasonable and radiologically correct?

3. Overall Support
Should the QA pair be retained as report-grounded supervision?

You must ONLY rely on the provided evidence.
Do NOT infer findings beyond the evidence.
--------------------------------------------------
[Evidence]
Finding:
{finding}

Impression:
{impression}

[Question]
{question}

[Answer]
{answer}
--------------------------------------------------
Scoring Criteria:
Factual Consistency:
0 = contradicted or absent from evidence
1 = weak or indirect evidence support
2 = partially supported but missing key qualifier
3 = directly supported with no hallucination

Clinical Validity:
0 = clinically implausible or radiologically wrong
1 = medically ambiguous or poorly grounded
2 = clinically plausible with minor imprecision
3 = radiologically coherent and clinically correct

Overall Support:
0 = discard
1 = manual review required
2 = usable but partially supported
3 = retain as fully report-grounded

Output ONLY valid JSON:
{{
  "factual_consistency": 0-3,
  "clinical_validity": 0-3,
  "overall_score": 0-3,
  "explanation": "brief explanation"
}}
\end{Verbatim}
\end{promptbox}

\section{Ethics, Privacy, and Data Governance}
\label{app:ethics}

\subsection{Ethics and Privacy}

This retrospective study and the use of clinical data for benchmark construction were approved through the responsible institutional ethics and data-governance procedures. All data handling follows the institutional data-governance protocol for retrospective medical research. Before benchmark construction, clinical images and reports are de-identified according to the approved protocol, including removal or masking of direct patient identifiers from reports and metadata. The public release plan follows the data-use permission of the source institution: source images, clinical reports, and other restricted artifacts will not be openly distributed.

\subsection{Template Design and Review}

The VQA templates are drafted from the structured breast MRI reporting schema and pathology attributes used in Stage I, then revised with clinical input to ensure that each template corresponds to a meaningful diagnostic question. Each template is associated with four fields: a clinical task category, a source structured key, an allowed answer space, and a question form. This binding prevents a template from asking for information that is absent from the extracted report evidence. Templates that are ambiguous, duplicate another task, or cannot be mapped to a single structured key are removed by clinical experts. The final templates are listed in Appendix~\ref{app:stage2_task} so that the mapping from task category to question form is explicit.

\subsection{Control of LLM-Generated Content}

The LLM is used for structured extraction and paraphrase proposal, but the accepted supervision is determined by report-derived structured fields and deterministic filters. For Stage I, outputs must be valid JSON under the predefined schema; unsupported or unmentioned attributes are set to \texttt{null}; values outside the allowed clinical vocabulary are discarded; and schema-invalid outputs are excluded from downstream question generation. For Stage III, paraphrasing is performed after the answer label has already been fixed by the structured field. A paraphrased pair is retained only if it preserves the same source key, answer label, and question type as the original template-generated pair. Thus, the LLM may change surface wording, but it cannot introduce a new label into the final dataset.

\subsection{Final VQA-Pair Filtering}

Each candidate VQA pair passes a final quality-control pipeline before inclusion. We remove pairs whose answer is unsupported by the source structured field, whose question refers to a missing or \texttt{null} attribute, whose answer is outside the template-specific vocabulary, or whose paraphrase changes the clinical slot being queried. For multiple-choice questions, we additionally filter answer leakage by matching the question text against the correct option and a synonym list for common clinical labels. We also remove duplicate, empty, contradictory, and malformed QA pairs. These steps are designed to reduce hallucinated labels, semantic drift during paraphrasing, and leakage from answer options into the question.

\section{Method Details}
\subsection{Greedy Prototype Selection Algorithm}
\label{app:gps_algorithm}
\begin{table*}[t]
\centering
\begin{tcolorbox}[
  enhanced,
  colback=gray!2,
  colframe=black!35,
  boxrule=0.5pt,
  arc=1mm,
  left=1.5mm,
  right=1.5mm,
  top=1mm,
  bottom=1mm
]
\small
\setlength{\tabcolsep}{3pt}
\begin{tabular}{@{}p{0.095\linewidth}p{0.845\linewidth}@{}}
\textbf{Input} &
Modality token matrix $X^{m} \in \mathbb{R}^{L_m \times d}$, prototype budget $k$, temperature $\tau_v$. \\
\textbf{Output} &
Prototype indices $S^{m,*}$ and position-aware prototypes $H^{m}$. \\
\end{tabular}
\vspace{0.35em}
\hrule
\vspace{0.45em}
\setlength{\tabcolsep}{3pt}
\begin{tabular}{@{}p{0.04\linewidth}p{0.19\linewidth}p{0.70\linewidth}@{}}
\textbf{1} & Normalize tokens &
$\tilde{X}^{m} \leftarrow \mathrm{L2Norm}(X^{m})$. \\
\textbf{2} & Compute affinity &
$A^{m} \leftarrow \tilde{X}^{m}(\tilde{X}^{m})^\top$, where $A^{m} \in \mathbb{R}^{L_m \times L_m}$. \\
\textbf{3} & Remove self-match &
$A^{m}_{ii} \leftarrow -\infty,\quad \forall i$. \\
\textbf{4} & Normalize rows &
$\displaystyle \hat{A}^{m}_{ij} \leftarrow
\frac{\exp(A^{m}_{ij}/\tau_v)}
{\sum_{j'=1}^{L_m}\exp(A^{m}_{ij'}/\tau_v)}$. \\
\textbf{5} & Initialize coverage &
$S \leftarrow \emptyset,\quad r_i \leftarrow -\infty$ for $i=1,\ldots,L_m$.
Here $r_i$ stores the best selected affinity for token $i$. \\

\textbf{6} & Greedy selection &
$\begin{aligned}[t]
&\text{for } t=1,\ldots,k:\\
&\qquad \mathrm{score}(s)=\frac{1}{L_m}\sum_{i=1}^{L_m}\max\!\left(r_i,\hat{A}^{m}_{is}\right),\\
&\qquad s\in\{1,\ldots,L_m\}\setminus S,\\
&\qquad s^\star \leftarrow \arg\max_s \mathrm{score}(s),\\
&\qquad S \leftarrow S \cup \{s^\star\},\\
&\qquad r_i \leftarrow \max\!\left(r_i,\hat{A}^{m}_{i s^\star}\right),\quad \forall i.
\end{aligned}$ \\
\textbf{7} & Form prototypes &
$S^{m,*}\leftarrow S,\quad P^{m} \leftarrow X^{m}[S^{m,*}]$. \\
\textbf{8} & Add position &
$H^{m} \leftarrow P^{m} + \mathrm{Embed}(\mathrm{pos}(S^{m,*}))$. \\
\textbf{9} & Return &
$S^{m,*},\ H^{m}$. \\
\end{tabular}
\end{tcolorbox}
\captionof{algorithm}{Greedy Prototype Selection (GPS)}\label{alg:grps}
\end{table*}
To identify a compact set of representative visual tokens for each modality, we adopt a greedy prototype selection strategy. 
Given the modality token matrix $X^{m}$, the algorithm iteratively selects $k$ tokens that maximize the overall affinity coverage of the token set. 
Specifically, we first normalize the tokens and compute a pairwise affinity matrix using cosine similarity. 
At each iteration, the candidate token that provides the largest marginal improvement in affinity coverage is selected as a new prototype. 
The detailed procedure is summarized in Algorithm~\ref{alg:grps}.

\subsection{Straight-Through Estimator for End-to-End Training}\label{alg:ste}
The greedy selection in Eq.~(\ref{equ3}) returns discrete token indices and is non-differentiable. However, the affinity matrix $\hat{A}$ is computed from upstream visual features that we want to train end-to-end with the downstream VQA objective. To propagate gradients through the selection step without changing its discrete behavior at inference, we apply a straight-through estimator (STE):
\begin{equation}
S^{m,*}_{\mathrm{ST}} = S^{m,*} + \hat{S}^{m} - \mathrm{stopgrad}(\hat{S}^{m}),
\label{eq:ste}
\end{equation}
where $\hat{S}^{m}$ is a \emph{soft prototype representation} computed as an affinity-weighted mixture of all tokens in $X^{m}$ using the normalized affinity scores $\hat{A}^{m}$, i.e., a differentiable surrogate of the hard selection.

\paragraph{Forward pass.} The first term $S^{m,*}$ dominates and equals the discrete greedy selection, so the model sees the same hard prototypes during both training and inference.

\paragraph{Backward pass.} The contribution of $S^{m,*}$ to the gradient is zero (it is a discrete index operation with no useful subgradient), and the contribution of $-\mathrm{stopgrad}(\hat{S}^{m})$ is also zero by construction. Gradients therefore flow only through the differentiable surrogate $\hat{S}^{m}$, which depends on the upstream visual features via $\hat{A}^{m}$.

\paragraph{What is trained vs.\ not trained.} The greedy index operation itself is \emph{not} trained—the set of selected indices is determined entirely by the current $\hat{A}^{m}$ at each forward pass. The trainable components are (i) the upstream 3D ViT that produces $X^{m}$ and (ii) any projection layers that shape the affinity space. Training therefore adjusts the feature space in which coverage is measured, which in turn changes which tokens get selected in subsequent forward passes.
\subsection{Self-Consistency Regularization Details}\label{alg:scrmu}
Self-Consistency Regularization with Marginal Utility is applied during task-specific fine-tuning to discourage degenerate always-on visual retrieval. The regularization compares the marginal utility of additional retrieved features and encourages the gate to remain active only when extra visual evidence reduces task loss. To stabilize optimization, the regularization is introduced after an initial supervised warm-up, so the model first learns stable VQA representations before the gating behavior is explicitly constrained.

\section{Implementation Details}
\label{app:implementation_details}

\subsection{Training Setup}
All experiments are conducted on a server with 8 NVIDIA H100 GPUs using BF16 mixed precision.
Unless otherwise stated, the main \dataset results use Qwen3-VL-4B as the backbone for \model-4B. The token-pruning efficiency comparison fixes the backbone to Qwen2.5-VL-3B for all pruning baselines and \model-3B, while the cross-model scaling study covers \model-3B (Qwen2.5-VL-3B), \model-4B (Qwen3-VL-4B), and \model-8B (Qwen3-VL-8B). All variants use the same two-stage training pipeline unless noted otherwise.

\subsection{Two-Stage Training Protocol}

\paragraph{Phase 1: Cross-modal representation alignment.}
The LLM backbone is kept \emph{frozen}; only the visual encoder, the GPS module, and the cross-modal projector are updated. This prevents catastrophic forgetting of language priors while the visual side learns domain-relevant representations.
For CT-based datasets (3D-RAD~\cite{gai20253drad}, DeepTumorVQA~\cite{chen2025deeptumor}), Phase 1 uses volume--caption pairs from CT-RATE~\cite{hamamci2024ct2rep}. For \dataset, Phase 1 trains the model to generate full radiology reports (findings and impressions) from multi-sequence MRI volumes, encouraging cross-modality alignment before downstream supervision.
Phase~1 hyperparameters are: learning rate $1\times10^{-4}$, effective batch size $32\times8{=}256$, 3 epochs, AdamW optimizer ($\beta_1{=}0.9$, $\beta_2{=}0.999$, $\varepsilon{=}10^{-8}$), cosine decay with a 5\% linear warm-up.

\paragraph{Phase 2: Task-specific fine-tuning.}
All parameters are unfrozen and the model is jointly fine-tuned on supervised VQA data with the standard language-modeling objective.
Phase~2 hyperparameters are: learning rate $8\times10^{-6}$ (swept from $5\times10^{-6}$ to $2\times10^{-5}$), effective batch size $32\times8{=}256$, 1 epoch, AdamW ($\beta_1{=}0.9$, $\beta_2{=}0.999$, $\varepsilon{=}10^{-8}$), weight decay $0.05$ (swept from $0.01$ to $0.1$; bias and layer-norm parameters excluded), cosine decay with a 3--5\% linear warm-up.
To stabilize joint optimization before the gating signal is active, the first 10\% of Phase~2 steps (0.1 epoch) run \emph{without} the Self-Consistency Regularization with Marginal Utility (Sec.~\ref{sec:scrmu}); the regularization is then enabled for the remainder of Phase~2 with weight $\beta{=}0.5$.
This staged activation lets the backbone first converge on the task loss, preventing degenerate early-stage gate saturation.

\subsection{Baseline Implementation Details}

\paragraph{Qwen3-VL (general MLLM baseline).}
We use the official implementation and pretrained weights. Multi-sequence MRI volumes are fed as \emph{native 3D video} inputs following the Qwen3-VL video encoding protocol, which treats temporal frames as volumetric slices and preserves their spatial structure. The model is fine-tuned on \dataset with the same two-stage pipeline and hyperparameters as \model.

\paragraph{Lingshu and other medical VLM baselines (HuLu-Med, OmniV).}
We use official implementations and pretrained weights. For Lingshu, each MRI volume is represented as a sequence of \emph{individual 2D images} (one per slice), so the number of input image tokens equals the number of slices in the volume; this multi-image input format aligns with Lingshu's native video-style API. All baselines are fine-tuned on \dataset following the same two-stage pipeline, and zero-shot variants are evaluated without any fine-tuning.

\paragraph{M3D (3D medical VLM baseline).}
We use the official implementation and pretrained weights. Because M3D accepts a single volumetric tensor, all available MRI modalities for a given case are \emph{concatenated along the channel dimension}; channels corresponding to missing modalities are filled with zeros. The 3D vision encoder is pre-trained on \dataset using the Phase~1 captioning objective, and the full model is then fine-tuned on \dataset VQA data, using the same data splits and optimizer settings as \model.

\paragraph{Visual token pruning baselines (VisionZip, DivPrune, MMTok).}
All three methods are reproduced from their official GitHub implementations and applied on top of \emph{our own fine-tuned Qwen2.5-VL-3B} checkpoint (trained with the two-stage protocol above, but without GPS/CaGA). Each method's token-selection module is inserted between the visual encoder output and the LLM input at inference time; no additional training is performed. Latency is measured on a single H100 GPU (batch size 1, BF16), averaged over 100 randomly sampled test examples from \dataset; CaGA retrieval overhead is included in \model's reported latency.

\paragraph{3D-specific models (Merlin, RadFM, CT-CHAT).}
Official checkpoints are evaluated in zero-shot mode on the CT benchmarks (DeepTumorVQA, 3D-RAD), following the original evaluation protocols without additional fine-tuning.

\begin{table*}[t]
\centering
\setlength{\tabcolsep}{2.6pt}
\renewcommand{\arraystretch}{0.9}
\resizebox{\linewidth}{!}{
\begin{tabular}{ll|cc|cc|cc|cc|cc|cc|cc|cc}
\toprule
\multirow{2}{*}{Setting} & \multirow{2}{*}{Base VLM} & \multicolumn{2}{c|}{GBA} & \multicolumn{2}{c|}{LDL} & \multicolumn{2}{c|}{Morph.} & \multicolumn{2}{c|}{MMFR} & \multicolumn{2}{c|}{LINR} & \multicolumn{2}{c|}{HD} & \multicolumn{2}{c|}{PP} & \multicolumn{2}{c}{Avg.} \\ \cline{3-18}
& & Acc. & BERT. & Acc. & BERT. & Acc. & BERT. & Acc. & BERT. & Acc. & BERT. & Acc. & BERT. & Acc. & BERT. & Acc. & BERT. \\ \midrule
\model-3B & Qwen2.5-VL-3B & 88.90 & 99.55 & 77.78 & 99.80 & 56.89 & 97.32 & 79.56 & 99.22 & 76.85 & 97.60 & 36.02 & 97.48 & 75.47 & 99.23 & 70.21 & 98.60 \\
\model-4B & Qwen3-VL-4B & 89.82 & 99.68 & 78.48 & 99.87 & 57.45 & 97.46 & 80.11 & 99.43 & 77.42 & 97.77 & 36.53 & 97.59 & 74.18 & 99.10 & 70.57 & 98.70 \\
\model-8B & Qwen3-VL-8B & 91.08 & 99.74 & 80.14 & 99.90 & 59.02 & 97.68 & 81.65 & 99.55 & 79.06 & 97.96 & 38.14 & 97.87 & 75.82 & 99.24 & 72.13 & 98.85 \\ \bottomrule
\end{tabular}
}
\caption{Full cross-model scaling on BreMRIs-VQA. Task abbreviations: GBA = Global Background 
Assessment, LDL = Lesion Detection and Localization, Morph. = Morphological 
Characterization, MMFR = Multi-Modal Functional Reasoning, LINR = Local 
Invasion and Nodal Reasoning, HD = Holistic Diagnosis, PP = Pathology 
Prediction. Avg. denotes the macro-average across the seven tasks.}
\label{tab:hyperparam_scaling_bremris_full}
\end{table*}

\section{Additional Experimental Results}
\label{app:add_results}

\subsection{Cross-Model Scaling on BreMRIs-VQA}
\label{app:cross-model-scaling}
Table~\ref{tab:hyperparam_scaling_bremris_full} reports the full task-wise results for the cross-model scaling study. \model-3B uses Qwen2.5-VL-3B, while \model-4B and \model-8B use Qwen3-VL-4B and Qwen3-VL-8B, respectively.

Performance improves monotonically across all seven tasks and both metrics as the backbone scales from 3B to 8B, demonstrating that the GPS/CaGA mechanism is backbone-agnostic and scales gracefully with LLM capacity. The 4B$\to$8B step (+1.56\,pp avg.\ Acc., +0.15\,pp BERTScore) yields the largest gains on Morphological Characterization and Multi-Modal Functional Reasoning — tasks requiring fine-grained pattern recognition across MRI sequences — suggesting that larger LLMs better leverage the multi-level evidence supplied by CaGA. The relative task ordering is preserved across all three scales, confirming that selective visual exposure provides consistent benefits independent of backbone capacity.

\subsection{Hyperparameter Sensitivity on BreMRIs-VQA}\label{app:abl_all}
Table~\ref{tab:hyperparam_sensitivity_bremris} sweeps four key hyperparameters one at a time while holding the others at their defaults (marked~$\dagger$). The model is robust throughout: all sweeps stay within $\pm0.7$\,pp of the default average accuracy.
\begin{table*}[t]
\centering
\setlength{\tabcolsep}{2.6pt}
\renewcommand{\arraystretch}{0.98}
\resizebox{0.98\linewidth}{!}{
\begin{tabular}{l|cc|cc|cc|cc|cc|cc|cc|cc}
\toprule
\multirow{2}{*}{Setting} & \multicolumn{2}{c|}{GBA} & \multicolumn{2}{c|}{LDL} & \multicolumn{2}{c|}{Morph.} & \multicolumn{2}{c|}{MMFR} & \multicolumn{2}{c|}{LINR} & \multicolumn{2}{c|}{HD} & \multicolumn{2}{c|}{PP} & \multicolumn{2}{c}{Avg.} \\ \cline{2-17}
& Acc. & BERT. & Acc. & BERT. & Acc. & BERT. & Acc. & BERT. & Acc. & BERT. & Acc. & BERT. & Acc. & BERT. & Acc. & BERT. \\ \midrule
\multicolumn{17}{c}{\textit{Effect of $\tau_v$ (visual-affinity temperature).}} \\ \midrule
$\tau_v$=0.05 & 89.10 & 99.55 & 77.64 & 99.75 & 56.72 & 97.21 & 79.42 & 99.21 & 76.83 & 97.53 & 36.05 & 97.32 & 73.26 & 98.92 & 69.86 & 98.50 \\
$\tau_v$=0.10$^\dagger$ & 89.82 & 99.68 & 78.48 & 99.87 & 57.45 & 97.46 & 80.11 & 99.43 & 77.42 & 97.77 & 36.53 & 97.59 & 74.18 & 99.10 & 70.57 & 98.70 \\
$\tau_v$=0.20 & 89.36 & 99.61 & 78.03 & 99.80 & 57.21 & 97.33 & 79.73 & 99.29 & 77.08 & 97.61 & 36.31 & 97.44 & 73.55 & 98.98 & 70.18 & 98.58 \\ \midrule
\multicolumn{17}{c}{\textit{Effect of $k$ (prototype budget).}} \\ \midrule
$k$=256 & 88.94 & 99.52 & 77.81 & 99.78 & 56.96 & 97.28 & 79.58 & 99.25 & 76.91 & 97.56 & 36.14 & 97.39 & 73.08 & 98.90 & 69.92 & 98.53 \\
$k$=512$^\dagger$ & 89.82 & 99.68 & 78.48 & 99.87 & 57.45 & 97.46 & 80.11 & 99.43 & 77.42 & 97.77 & 36.53 & 97.59 & 74.18 & 99.10 & 70.57 & 98.70 \\
$k$=1024 & 89.41 & 99.60 & 78.15 & 99.82 & 57.02 & 97.30 & 79.66 & 99.31 & 76.98 & 97.60 & 36.36 & 97.47 & 73.04 & 98.94 & 70.09 & 98.58 \\ \midrule
\multicolumn{17}{c}{\textit{Effect of $s$ (gate sharpness).}} \\ \midrule
$s$=5 & 89.28 & 99.58 & 77.92 & 99.79 & 57.18 & 97.35 & 79.51 & 99.24 & 76.95 & 97.58 & 36.22 & 97.42 & 73.18 & 98.93 & 70.03 & 98.56 \\
$s$=10$^\dagger$ & 89.82 & 99.68 & 78.48 & 99.87 & 57.45 & 97.46 & 80.11 & 99.43 & 77.42 & 97.77 & 36.53 & 97.59 & 74.18 & 99.10 & 70.57 & 98.70 \\
$s$=20 & 89.47 & 99.62 & 78.11 & 99.81 & 57.24 & 97.37 & 79.88 & 99.33 & 77.16 & 97.65 & 36.41 & 97.50 & 73.22 & 98.96 & 70.21 & 98.61 \\ \midrule
\multicolumn{17}{c}{\textit{Effect of $\beta$ (marginal-utility regularization weight).}} \\ \midrule
$\beta$=0.1 & 89.33 & 99.59 & 78.06 & 99.81 & 57.11 & 97.34 & 79.74 & 99.30 & 77.05 & 97.62 & 36.27 & 97.45 & 73.28 & 98.95 & 70.12 & 98.58 \\
$\beta$=0.5$^\dagger$ & 89.82 & 99.68 & 78.48 & 99.87 & 57.45 & 97.46 & 80.11 & 99.43 & 77.42 & 97.77 & 36.53 & 97.59 & 74.18 & 99.10 & 70.57 & 98.70 \\
$\beta$=1.0 & 89.08 & 99.54 & 77.77 & 99.77 & 56.98 & 97.25 & 79.53 & 99.24 & 76.89 & 97.55 & 36.16 & 97.38 & 73.17 & 98.91 & 69.94 & 98.52 \\ \bottomrule
\end{tabular}
}
\caption{Hyperparameter sensitivity on \dataset. Each block varies one hyperparameter while keeping the others fixed at their default values, marked with $\dagger$. We report multiple-choice accuracy (Acc.) and free-text BERTScore (BERT). GBA = Global Background 
Assessment, LDL = Lesion Detection and Localization, Morph. = Morphological 
Characterization, MMFR = Multi-Modal Functional Reasoning, LINR = Local 
Invasion and Nodal Reasoning, HD = Holistic Diagnosis, PP = Pathology 
Prediction. }
\label{tab:hyperparam_sensitivity_bremris}
\end{table*}
For visual-affinity temperature $\tau_v$, the optimal 0.10 balances prototype selectivity: lower values concentrate affinities too aggressively (risking coverage collapse on diverse multi-sequence inputs), while higher values over-smooth inter-prototype similarities and reduce discriminability. The prototype budget $k{=}512$ maximizes coverage without introducing redundant prototypes that dilute the selection signal, consistent with the latency--accuracy trade-off in Sec.~\ref{sec:efficiency}. Gate sharpness $s{=}10$ provides adequate binary resolution without causing premature gate saturation. For the regularization weight $\beta$, a moderate 0.5 penalizes degenerate always-on retrieval without over-suppressing the gating mechanism; both $\beta{=}0.1$ (too lenient) and $\beta{=}1.0$ (too aggressive) degrade average accuracy by 0.4--0.6\,pp.

\subsection{Full DeepTumorVQA Results}\label{app:deeptumorvqa_full}
Table~\ref{tab:deeptumorvqa_full} reports the complete subtype-level results on DeepTumorVQA. Subtypes marked with $\ast$ are evaluated using MRA on free-text numerical answers.
\begin{table*}[t]
\centering
\setlength{\tabcolsep}{2.2pt}
\renewcommand{\arraystretch}{0.92}
\resizebox{\linewidth}{!}{
\begin{tabular}{c|c|cccccc|cccccc}
\toprule
\multirow{3}{*}{Type} & \multirow{3}{*}{Subtype} & \multicolumn{6}{c|}{Multiple-choice} & \multicolumn{6}{c}{Free-text} \\ \cline{3-14}
 &  & Merlin & M3D-L2 & M3D-P3 & CT-CHAT & RadFM & \model & Merlin & M3D-L2 & M3D-P3 & CT-CHAT & RadFM & \model \\ \cline{3-14}
 &  & 7B & 7B & 4B & 7B & 13B & 4B & 7B & 7B & 4B & 7B & 13B & 4B \\ \midrule
\multirow{4}{*}{Meas} & Lesion Volume$\ast$ & 0.253 & 0.815 & 0.825 & 0.833 & 0.815 & \cellcolor{blue!3}0.819 & 0.079 & 0.085 & 0.079 & 0.075 & 0.112 & \cellcolor{blue!3}0.258 \\
 & Organ HU$\ast$ & 0.254 & 0.638 & 0.640 & 0.637 & 0.647 & \cellcolor{blue!3}0.637 & 0.487 & 0.490 & 0.491 & 0.513 & 0.608 & \cellcolor{blue!3}0.620 \\
 & Organ Volume$\ast$ & 0.262 & 0.747 & 0.754 & 0.750 & 0.755 & \cellcolor{blue!3}0.748 & 0.526 & 0.535 & 0.528 & 0.549 & 0.583 & \cellcolor{blue!3}0.685 \\
 & \textbf{Average} & 0.256 & 0.733 & 0.740 & \textbf{0.740} & \underline{0.739} & \cellcolor{blue!3}0.735 & 0.364 & 0.370 & 0.366 & 0.379 & \underline{0.434} & \cellcolor{blue!3}\textbf{0.521}  \\ \hline
\multirow{7}{*}{Recog} & Colon Lesion & 0.859 & 0.859 & 0.859 & 0.859 & 0.856 & \cellcolor{blue!3}0.859 & 0.859 & 0.859 & 0.859 & 0.859 & 0.893 & \cellcolor{blue!3}0.859 \\
 & Kidney Cyst & 0.797 & 0.797 & 0.797 & 0.797 & 0.861 & \cellcolor{blue!3}0.856 & 0.797 & 0.797 & 0.797 & 0.797 & 0.864 & \cellcolor{blue!3}0.856 \\
 & Kidney Lesion & 0.495 & 0.510 & 0.501 & 0.514 & 0.668 & \cellcolor{blue!3}0.491 & 0.511 & 0.515 & 0.490 & 0.507 & 0.692 & \cellcolor{blue!3}0.520 \\
 & Kidney Tumor & 0.564 & 0.574 & 0.574 & 0.574 & 0.886 & \cellcolor{blue!3}0.852 & 0.574 & 0.574 & 0.574 & 0.574 & 0.890 & \cellcolor{blue!3}0.852 \\
 & Liver Lesion & 0.535 & 0.524 & 0.517 & 0.524 & 0.652 & \cellcolor{blue!3}0.516 & 0.524 & 0.524 & 0.524 & 0.524 & 0.662 & \cellcolor{blue!3}0.526 \\
 & Pancreatic Lesion & 0.718 & 0.718 & 0.718 & 0.718 & 0.810 & \cellcolor{blue!3}0.669 & 0.718 & 0.718 & 0.718 & 0.718 & 0.871 & \cellcolor{blue!3}0.675 \\
 & \textbf{Average} & 0.661 & 0.664 & 0.661 & 0.664 & \textbf{0.789} & \cellcolor{blue!3}\underline{0.707} & 0.664 & 0.665 & 0.660 & 0.663 & \textbf{0.812} & \cellcolor{blue!3}\underline{0.715} \\ \hline
\multirow{15}{*}{VisRsn} & Adj Organ & 0.217 & 0.565 & 0.609 & 0.609 & 0.609 & \cellcolor{blue!3}0.652 & 0.174 & 0.174 & 0.304 & 0.304 & 0.435 & \cellcolor{blue!3}0.044 \\
 & Inter-Segment & 0.470 & 0.567 & 0.576 & 0.572 & 0.591 & \cellcolor{blue!3}0.714 & 0.577 & 0.561 & 0.592 & 0.589 & 0.456 & \cellcolor{blue!3}0.524 \\
 & Kidney Vol Comp & 0.347 & 0.370 & 0.364 & 0.372 & 0.386 & \cellcolor{blue!3}0.414 & 0.350 & 0.370 & 0.356 & 0.370 & 0.386 & \cellcolor{blue!3}0.359 \\
 & Lesion Attenuation & 0.317 & 0.541 & 0.539 & 0.544 & 0.555 & \cellcolor{blue!3}0.679 & 0.526 & 0.544 & 0.548 & 0.542 & 0.521 & \cellcolor{blue!3}0.505 \\
 & Lesion Diameter$\ast$ & 0.263 & 0.778 & 0.783 & 0.781 & 0.766 & \cellcolor{blue!3}0.725 & 0.182 & 0.209 & 0.233 & 0.269 & 0.232 & \cellcolor{blue!3}0.414 \\
 & Lesion Location & 0.307 & 0.310 & 0.310 & 0.340 & 0.340 & \cellcolor{blue!3}0.476 & 0.359 & 0.353 & 0.337 & 0.353 & 0.334 & \cellcolor{blue!3}0.133 \\
 & Lesion Slice$\ast$ & 0.241 & 0.672 & 0.684 & 0.672 & 0.664 & \cellcolor{blue!3}0.684 & 0.524 & 0.533 & 0.510 & 0.513 & 0.672 & \cellcolor{blue!3}0.696 \\
 & Lesion Count \& Loc$\ast$ & 0.583 & 0.861 & 0.860 & 0.862 & 0.861 & \cellcolor{blue!3}0.856 & 0.534 & 0.534 & 0.534 & 0.534 & 0.506 & \cellcolor{blue!3}0.616  \\
 & Lesion Count$\ast$ & 0.455 & 0.781 & 0.784 & 0.796 & 0.790 & \cellcolor{blue!3}0.982 & 0.000 & 0.000 & 0.000 & 0.000 & 0.001 & \cellcolor{blue!3}0.914 \\
 & Lesion Outlier & 0.521 & 0.507 & 0.549 & 0.451 & 0.493 & \cellcolor{blue!3}0.423 & 0.451 & 0.535 & 0.535 & 0.577 & 0.521 & \cellcolor{blue!3}0.549 \\
 & Liver Lesion Clust & 0.331 & 0.438 & 0.475 & 0.463 & 0.469 & \cellcolor{blue!3}0.669 & 0.388 & 0.469 & 0.469 & 0.431 & 0.513 & \cellcolor{blue!3}0.413 \\
 & Organ Aggrega.$\ast$ & 0.257 & 0.660 & 0.667 & 0.655 & 0.661 & \cellcolor{blue!3}0.635 & 0.577 & 0.569 & 0.586 & 0.574 & 0.621 & \cellcolor{blue!3}0.720 \\
 & Organ Enlarge. & 0.736 & 0.736 & 0.736 & 0.736 & 0.746 & \cellcolor{blue!3}0.717 & 0.736 & 0.736 & 0.736 & 0.736 & 0.759 & \cellcolor{blue!3}0.713 \\
 & Tumor-Organ HU$\ast$ & 0.296 & 0.836 & 0.839 & 0.821 & 0.821 & \cellcolor{blue!3}0.802 & 0.113 & 0.122 & 0.139 & 0.197 & 0.189 & \cellcolor{blue!3}0.384 \\
 & \textbf{Average} & 0.382 & 0.616 & 0.627 & 0.620 & \underline{0.625} & \cellcolor{blue!3}\textbf{0.673} & 0.392 & 0.408 & 0.420 & 0.428 & \underline{0.439} & \cellcolor{blue!3}\textbf{0.499}  \\ \hline
\multirow{7}{*}{MedRsn} & Fatty Liver & 0.318 & 0.461 & 0.455 & 0.481 & 0.481 & \cellcolor{blue!3}0.558 & 0.481 & 0.481 & 0.396 & 0.487 & 0.578 & \cellcolor{blue!3}0.422 \\
 & Lesion Type & 0.865 & 0.865 & 0.865 & 0.865 & 0.865 & \cellcolor{blue!3}0.865 & 0.865 & 0.865 & 0.865 & 0.865 & 0.851 & \cellcolor{blue!3}0.865 \\
 & Cyst Resecta. & 0.371 & 0.657 & 0.800 & 0.800 & 0.771 & \cellcolor{blue!3}0.771 & 0.800 & 0.800 & 0.800 & 0.800 & 0.771 & \cellcolor{blue!3}0.829 \\
 & Lesion Resectab. & 0.379 & 0.483 & 0.483 & 0.483 & 0.483 & \cellcolor{blue!3}1.000 & 0.414 & 0.483 & 0.483 & 0.483 & 0.483 & \cellcolor{blue!3}1.000 \\
 & Pancrea. Steat. & 0.526 & 0.526 & 0.513 & 0.513 & 0.579 & \cellcolor{blue!3}0.526 & 0.526 & 0.526 & 0.526 & 0.526 & 0.658 & \cellcolor{blue!3}0.566 \\
 & Tumor Staging & 0.216 & 0.351 & 0.243 & 0.189 & 0.324 & \cellcolor{blue!3}0.216 & 0.216 & 0.216 & 0.297 & 0.135 & 0.432 & \cellcolor{blue!3}0.297 \\
 & \textbf{Average} & 0.446 & 0.557 & 0.560 & 0.555 & \underline{0.584} & \cellcolor{blue!3}\textbf{0.656} & 0.550 & 0.562 & 0.561 & 0.549 & \underline{0.629} & \cellcolor{blue!3}\textbf{0.663}  \\ \hline
Total & \textbf{Average} & 0.440 & 0.626 & 0.632 & 0.628 & \underline{0.662} & \cellcolor{blue!3}\textbf{0.687} & 0.478 & 0.489 & 0.493 & 0.497 & \underline{0.555} & \cellcolor{blue!3}\textbf{0.599}  \\ \bottomrule
\end{tabular}}
\caption{Full subtype-level performance on DeepTumorVQA. Subtypes marked with $\ast$ indicate free-text numerical answers evaluated using MRA (higher is better).}
\label{tab:deeptumorvqa_full}
\end{table*}
\model-4B achieves the highest overall average in both multiple-choice (0.687) and free-text (0.599), outperforming even the 13B RadFM. The largest margins appear in Visual Reasoning (VisRsn: +0.048\,MC, +0.060\,FT over the second-best) and Medical Reasoning (MedRsn: +0.072\,MC, +0.034\,FT), task categories that require integrating cross-slice spatial cues — precisely the regime where CaGA's on-demand evidence retrieval is most effective. Notably, for Lesion Count, \model reaches 0.982 accuracy in MC and 0.914 in FT, far ahead of all baselines, indicating that selective prototype coverage preserves the counting evidence that dense token exposure dilutes. In Measurement subtypes (Meas), \model leads on free-text MRA (0.521 vs.\ RadFM's 0.434) despite trailing in MC ($-0.030$ vs.\ CT-CHAT), suggesting that the generative pathway benefits more from focused visual access than the classification head. In Recognition (Recog), the larger RadFM (13B) leads at 0.789/0.812; \model-4B reaches 0.707/0.715, competitive given the 3$\times$ parameter gap.

\subsection{Full 3D-RAD Ablation Results}
\label{abl:3drad}
Table~\ref{tab:ablation_3drad_tokens_full} reports the full per-task ablation on 3D-RAD, extending the summary in the main text. E.D.\ = Existence Detection, S.T.D.\ = Static Temporal Diagnosis, L.T.D.\ = Longitudinal Temporal Diagnosis.
\begin{table*}[!htbp]
\centering
\setlength{\tabcolsep}{3.8pt}
\renewcommand{\arraystretch}{0.92}
\resizebox{\linewidth}{!}{
\begin{tabular}{c|ccc|ccc|ccc|ccc|c}
\toprule
\multirow{2}{*}{Metrics} & E.D. & S.T.D. & L.T.D. & \multicolumn{3}{c|}{Medical Measurement} & \multicolumn{3}{c|}{Image Observation} & \multicolumn{3}{c|}{Anomaly Detection} & \multirow{2}{*}{\begin{tabular}[c]{@{}c@{}}Exposed \\ tokens\end{tabular}} \\ \cline{2-13}
 & Acc. & Acc. & Acc. & BLEU & ROUGE & BERT & BLEU & ROUGE & BERT & BLEU & ROUGE & BERT &  \\ \midrule
Baseline Zs. & 3.52 & 0 & 0.15 & 0.11 & 0.08 & 76.65 & 0.26 & 0.44 & 76.22 & 0.24 & 0.54 & 76.65 & 4608 \\
\model Phase1 & 22.94 & 10.54 & 21.12 & 10.47 & 20.78 & 91.23 & 4.74 & 17.94 & 86.5 & 5.36 & 17.28 & 86.24 & 512 \\
Baseline Ft. & 82.01 & 45.16 & 72.19 & 32.95 & 33.78 & 95.03 & 42.63 & 45.25 & 91.92 & 34.27 & 37.54 & 90.58 & 4608 \\
w/o CaGA & 75.73 & 41.70 & 66.66 & 30.43 & 31.19 & 87.75 & 39.36 & 41.78 & 84.88 & 31.64 & 34.66 & 83.64 & 512 \\
$k$=256 & 82.11 & 51.36 & 75.63 & 37.21 & 39.56 & 95.95 & 48.84 & 53.72 & 93.10 & 39.07 & 44.85 & 91.56 & 256 \\
$k$=1024 & 82.37 & 51.11 & 73.39 & 33.94 & 35.68 & 95.49 & 48.72 & 53.43 & 93.06 & 38.97 & 43.18 & 91.58 & 1024 \\
$k$=2048 & 82.36 & 52.56 & 76.87 & 34.73 & 36.79 & 95.78 & 43.62 & 47.27 & 92.33 & 38.80 & 42.57 & 91.42 & 2048 \\
w/o Regul. & 81.44 & 51.45 & 74.62 & 36.53 & 38.78 & 95.00 & 48.51 & 53.88 & 92.19 & 39.60 & 45.36 & 90.69 & 512 \\
\rowcolor{blue!3}\model & 82.26 & 51.77 & 75.28 & 36.78 & 39.01 & 95.93 & 48.81 & 54.16 & 93.09 & 39.83 & 45.58 & 91.58 & 512 \\ \bottomrule
\end{tabular}
}
\caption{Full ablation study on 3D-RAD. We report task performance and the average number of exposed input visual tokens. E.D. = Existence Detection, S.T.D. = Static Temporal Diagnosis, L.T.D. = Longitudinal Temporal Diagnosis, Zs. = Zero-shot, Ft. = Fine-tune, Acc. = Accuracy, and BERT = BERTScore.}\label{tab:ablation_3drad_tokens_full}
\end{table*}
Removing CaGA (w/o CaGA) incurs the largest single drop on Longitudinal Temporal Diagnosis (L.T.D.: 66.66 vs.\ 75.28 for the full model) and Static Temporal Diagnosis (S.T.D.: 41.70 vs.\ 51.77), where multi-slice evidence retrieval is essential for identifying temporal changes — confirming that prototype compression alone is insufficient for temporally demanding tasks. The $k$ sweep shows that $k{=}512$ achieves the best aggregate: $k{=}256$ degrades free-text metrics despite maintaining competitive E.D.\ accuracy, as fewer prototypes suppress multi-scale cues needed for generation; $k{\geq}1024$ saturates accuracy while consuming more tokens with no BERTScore benefit. Removing the marginal-utility regularization (w/o Regul.) degrades E.D.\ accuracy by 0.82\,pp and visibly hurts free-text BLEU/ROUGE across all three generative task groups, confirming that the regularization suppresses degenerate always-on gate patterns that dilute generation quality. The Phase~1-only checkpoint (without fine-tuning) shows that representation alignment through captioning is a necessary precursor, yielding non-trivial free-text quality (BERTScore~$\approx$~86--91) but near-zero discriminative accuracy, which supervision subsequently converts into strong task-specific performance.

\begin{figure*}[t]
    \centering
    \begin{subfigure}[b]{0.32\linewidth}
        \centering
        \includegraphics[width=\linewidth]{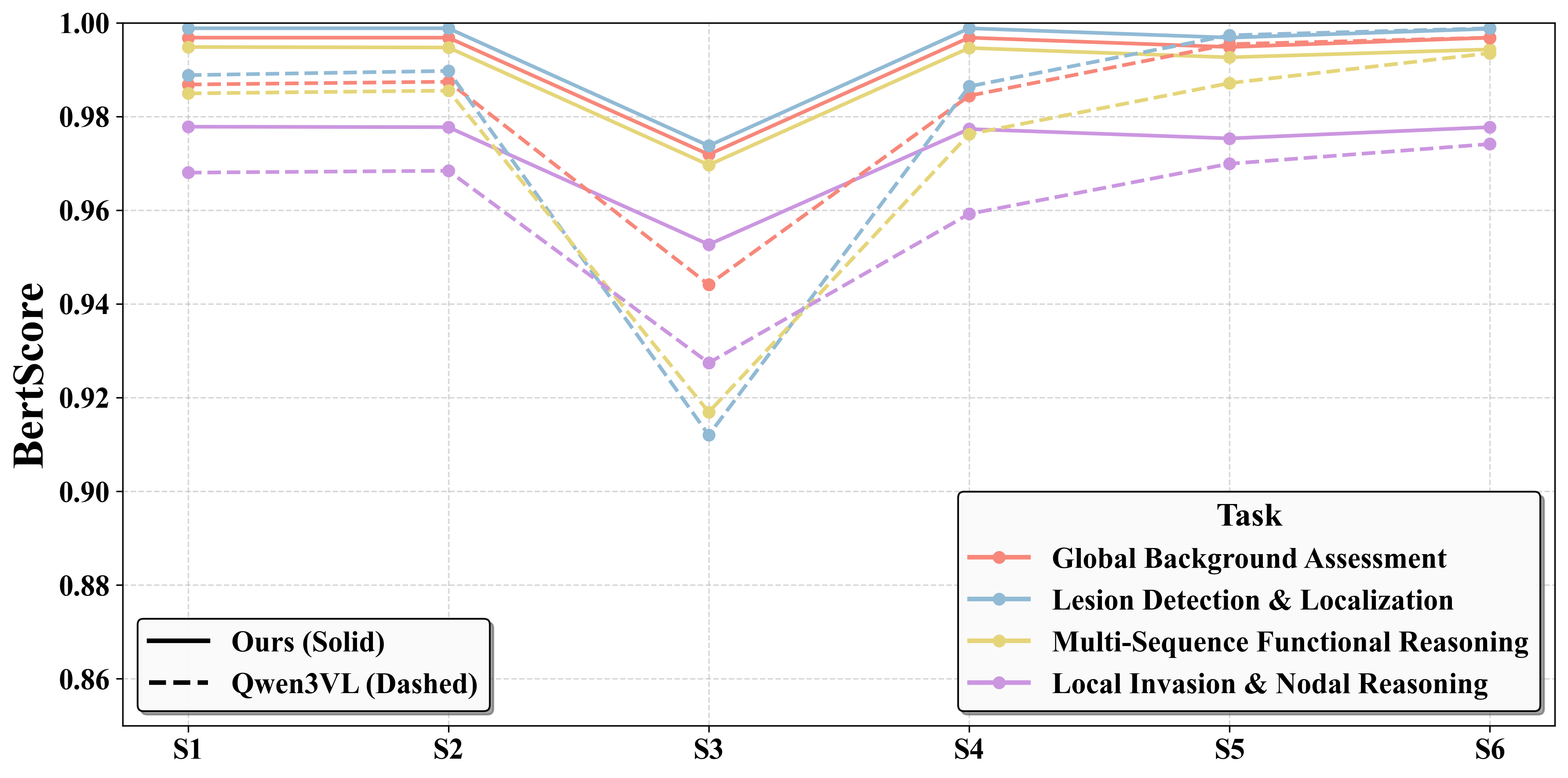}
        \caption{Free-text BERTScore.}
        \label{fig:modality_ablation_ft}
    \end{subfigure}
    \hfill
    \begin{subfigure}[b]{0.32\linewidth}
        \centering
        \includegraphics[width=\linewidth]{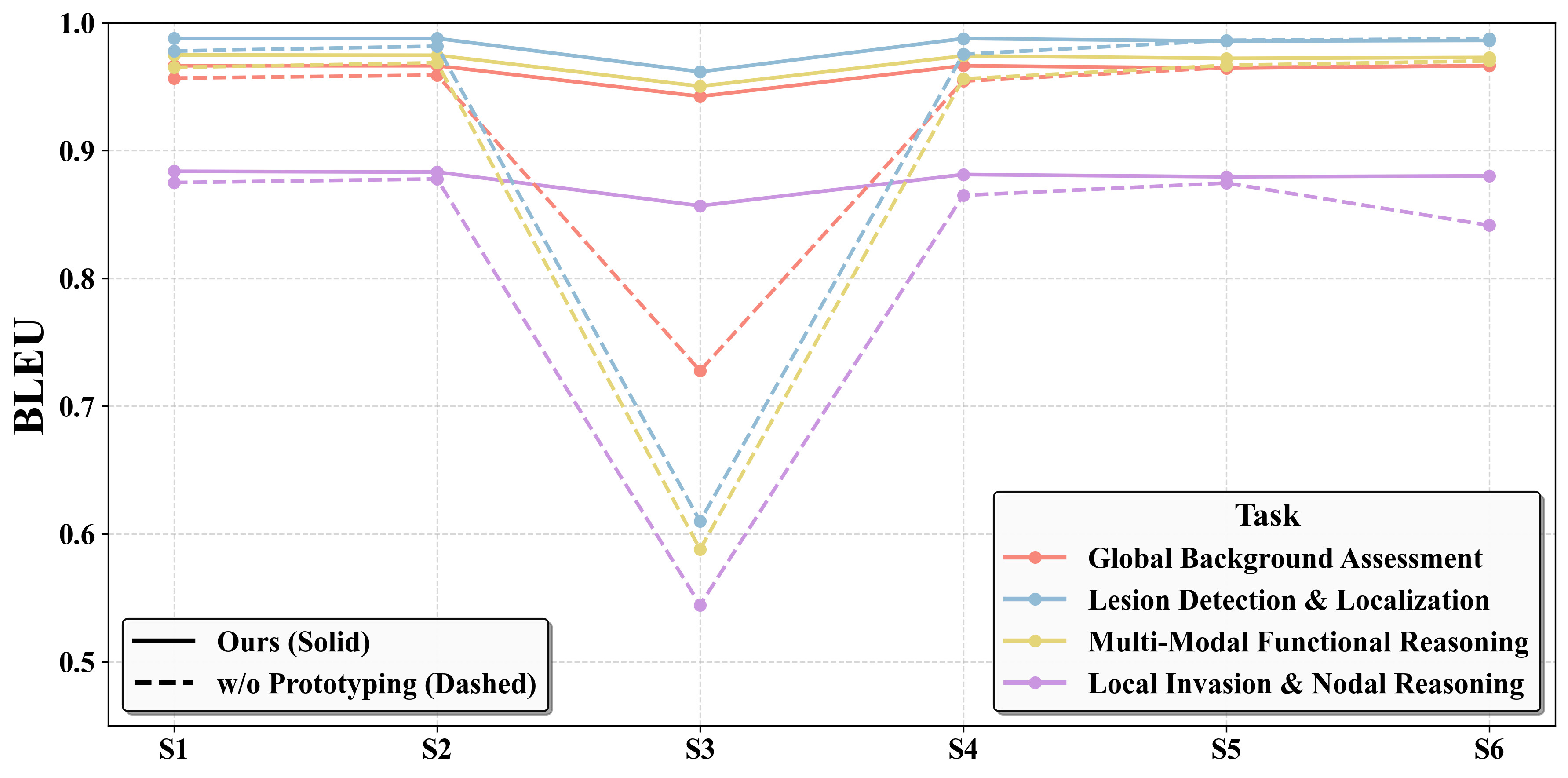}
        \caption{Free-text BLEU.}
        \label{fig:modality_ablation_bleu}
    \end{subfigure}
    \hfill
    \begin{subfigure}[b]{0.32\linewidth}
        \centering
        \includegraphics[width=\linewidth]{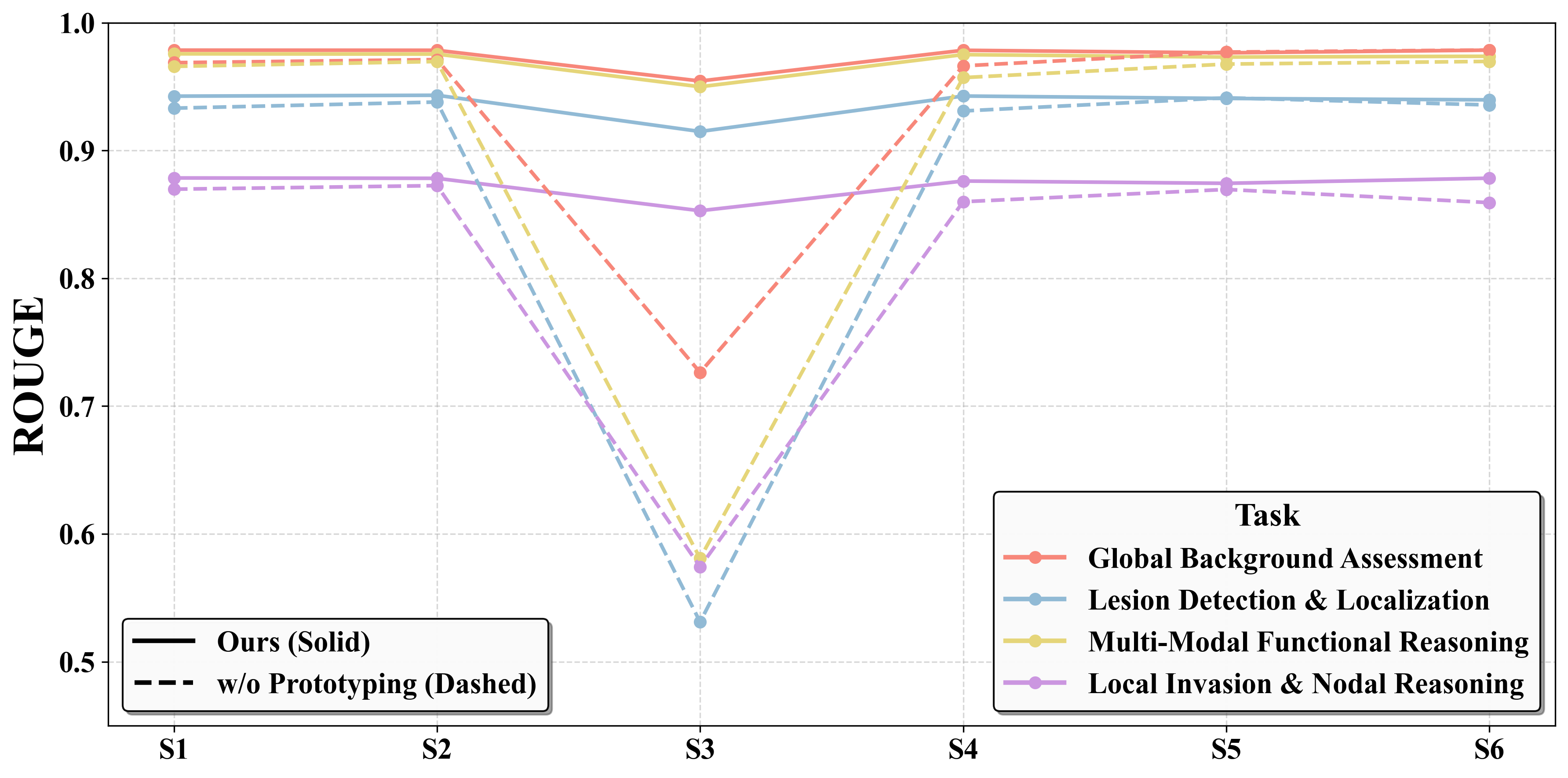}
        \caption{Free-text ROUGE.}
        \label{fig:modality_ablation_rouge}
    \end{subfigure}
    \vspace{-0.5em}
    \caption{\textbf{Robustness to modality availability under different settings.}
    \model (solid) vs.\ w/o selective exposure (dashed) under six modality settings (S1--S6).
    \textbf{(a)} BERTScore, \textbf{(b)} BLEU, \textbf{(c)} ROUGE for free-text quality.}
    \label{fig:modality_addition}
\end{figure*}
\subsection{Additional Results for Modality Settings}\label{app:add_modality_setting}
Table~\ref{tab:modality_settings_full} provides the complete numerical breakdown of modality-robustness results across all six settings and seven clinical tasks.
The six settings correspond to progressively richer combinations of breast MRI sequences: \textit{S1} uses only T1-weighted imaging (anatomical structure); \textit{S2} adds dynamic contrast-enhanced sequences (DCE) for vascular enhancement patterns; \textit{S3} uses diffusion-weighted imaging (DWI), apparent diffusion coefficient (ADC), and T2-weighted imaging for signal characterization; \textit{S4} combines T1w, DCE, T2w, and DWI; \textit{S5} further adds T1-STIR to S4; and \textit{S6} uses all available sequences.
Across every task and metric, \model maintains consistently smaller performance gaps relative to S6 compared with the w/o Proto.\ baseline, confirming that GPS prototype coverage and CaGA on-demand retrieval collectively reduce over-reliance on any single modality.

\begin{table*}[!htbp]
\centering
\setlength{\tabcolsep}{3pt}
\renewcommand{\arraystretch}{0.88}

\resizebox{\linewidth}{!}{\begin{tabular}{ll|cccccc|cccccc}
\toprule
\multirow{2}{*}{\textbf{Task}} & \multirow{2}{*}{\textbf{Model}} &
  \multicolumn{6}{c|}{\textbf{Acc.}} &
  \multicolumn{6}{c}{\textbf{BLEU}} \\
 & & \textbf{S1} & \textbf{S2} & \textbf{S3} & \textbf{S4} & \textbf{S5} & \textbf{S6} &
     \textbf{S1} & \textbf{S2} & \textbf{S3} & \textbf{S4} & \textbf{S5} & \textbf{S6} \\
\midrule
Global Background Assessment
  & w/o Proto. & 89.82 & 89.82 & 83.10 & 87.78 & 79.97 & 89.43 & 95.66 & 95.90 & 72.76 & 95.43 & 96.49 & 96.63 \\
  & SeVeR      & \textbf{89.82} & \textbf{89.82} & \textbf{89.82} & \textbf{87.78} & \textbf{80.77} & \textbf{89.82} & \textbf{96.63} & \textbf{96.63} & \textbf{96.65} & \textbf{96.63} & \textbf{96.44} & \textbf{96.63} \\ \midrule
Lesion Detection \& Localization
  & w/o Proto. & 78.39 & 76.60 & 69.76 & 72.68 & 54.69 & 73.55 & 97.78 & 98.16 & 60.99 & 97.53 & 98.62 & 98.74 \\
  & SeVeR      & \textbf{78.39} & \textbf{76.60} & \textbf{80.21} & \textbf{72.67} & \textbf{54.80} & \textbf{78.48} & \textbf{98.77} & \textbf{98.77} & \textbf{98.62} & \textbf{98.76} & \textbf{98.56} & \textbf{98.61} \\ \midrule
Morphological Characterization
  & w/o Proto. & 57.35 & 49.50 & 37.80 & 53.65 & 27.92 & 57.68 & 73.63 & 74.42 & 41.22 & 73.96 & 74.78 & 70.34 \\
  & SeVeR      & \textbf{57.35} & \textbf{49.50} & \textbf{37.73} & \textbf{53.65} & \textbf{28.04} & \textbf{57.45} & \textbf{74.38} & \textbf{74.83} & \textbf{72.30} & \textbf{74.89} & \textbf{74.74} & \textbf{73.46} \\ \midrule
Multi-Modal Functional Reasoning
  & w/o Proto. & 78.75 & 73.60 & 61.43 & 64.70 & 51.33 & 79.81 & 96.50 & 96.86 & 58.80 & 95.60 & 96.66 & 97.01 \\
  & SeVeR      & \textbf{78.75} & \textbf{73.60} & \textbf{62.75} & \textbf{64.70} & \textbf{55.95} & \textbf{80.11} & \textbf{97.47} & \textbf{97.45} & \textbf{97.46} & \textbf{97.40} & \textbf{97.21} & \textbf{97.28} \\ \midrule
Local Invasion \& Nodal Reasoning
  & w/o Proto. & 77.01 & 76.96 & 67.05 & 69.64 & 66.74 & 78.54 & 87.49 & 87.77 & 54.44 & 86.49 & 87.45 & 84.14 \\
  & SeVeR      & \textbf{77.01} & \textbf{76.96} & \textbf{77.99} & \textbf{69.64} & \textbf{68.25} & \textbf{77.42} & \textbf{88.37} & \textbf{88.31} & \textbf{87.87} & \textbf{88.12} & \textbf{87.94} & \textbf{88.01} \\ \midrule
Holistic Diagnosis
  & w/o Proto. & 34.97 & 37.08 & 22.26 & 37.50 & 30.66 & 35.86 & 87.44 & 87.88 & 43.93 & 86.56 & 87.53 & 87.35 \\
  & SeVeR      & \textbf{34.97} & \textbf{37.08} & \textbf{22.82} & \textbf{37.50} & \textbf{31.75} & \textbf{36.53} & \textbf{88.32} & \textbf{88.32} & \textbf{88.02} & \textbf{88.29} & \textbf{88.11} & \textbf{88.22} \\ \midrule
Pathology Prediction
  & w/o Proto. & 72.79 & 68.07 & 63.18 & 69.75 & 63.75 & 71.29 & 94.39 & 94.78 & 61.47 & 93.48 & 94.52 & 94.89 \\
  & SeVeR      & \textbf{72.79} & \textbf{68.07} & \textbf{61.29} & \textbf{69.75} & \textbf{65.25} & \textbf{74.18} & \textbf{95.34} & \textbf{95.39} & \textbf{95.70} & \textbf{95.34} & \textbf{95.15} & \textbf{95.34} \\
\bottomrule
\end{tabular}}

\smallskip
\resizebox{\linewidth}{!}{\begin{tabular}{ll|cccccc|cccccc}
\toprule
\multirow{2}{*}{\textbf{Task}} & \multirow{2}{*}{\textbf{Model}} &
  \multicolumn{6}{c|}{\textbf{ROUGE}} &
  \multicolumn{6}{c}{\textbf{BERTScore}} \\
 & & \textbf{S1} & \textbf{S2} & \textbf{S3} & \textbf{S4} & \textbf{S5} & \textbf{S6} &
     \textbf{S1} & \textbf{S2} & \textbf{S3} & \textbf{S4} & \textbf{S5} & \textbf{S6} \\
\midrule
Global Background Assessment
  & w/o Proto. & 96.87 & 97.11 & 72.63 & 96.62 & 97.70 & 97.86 & 98.68 & 98.74 & 94.41 & 98.44 & 99.54 & 99.68 \\
  & SeVeR      & \textbf{97.85} & \textbf{97.84} & \textbf{97.88} & \textbf{97.84} & \textbf{97.64} & \textbf{97.85} & \textbf{99.68} & \textbf{99.68} & \textbf{99.68} & \textbf{99.68} & \textbf{99.48} & \textbf{99.68} \\ \midrule
Lesion Detection \& Localization
  & w/o Proto. & 93.31 & 93.79 & 53.11 & 93.09 & 94.12 & 93.56 & 98.88 & 98.97 & 91.20 & 98.64 & 99.73 & 99.88 \\
  & SeVeR      & \textbf{94.25} & \textbf{94.32} & \textbf{93.83} & \textbf{94.26} & \textbf{94.07} & \textbf{93.96} & \textbf{99.88} & \textbf{99.88} & \textbf{99.87} & \textbf{99.88} & \textbf{99.68} & \textbf{99.87} \\ \midrule
Morphological Characterization
  & w/o Proto. & 71.98 & 72.43 & 47.23 & 71.50 & 72.30 & 71.92 & 96.51 & 96.58 & 91.80 & 95.69 & 96.75 & 97.24 \\
  & SeVeR      & \textbf{72.71} & \textbf{72.90} & \textbf{72.64} & \textbf{72.85} & \textbf{72.70} & \textbf{72.63} & \textbf{97.48} & \textbf{97.50} & \textbf{97.47} & \textbf{97.49} & \textbf{97.30} & \textbf{97.46} \\ \midrule
Multi-Modal Functional Reasoning
  & w/o Proto. & 96.58 & 96.96 & 58.09 & 95.70 & 96.76 & 96.97 & 98.49 & 98.55 & 91.69 & 97.62 & 98.71 & 99.35 \\
  & SeVeR      & \textbf{97.56} & \textbf{97.54} & \textbf{97.42} & \textbf{97.50} & \textbf{97.31} & \textbf{97.37} & \textbf{99.48} & \textbf{99.47} & \textbf{99.45} & \textbf{99.46} & \textbf{99.26} & \textbf{99.43} \\ \midrule
Local Invasion \& Nodal Reasoning
  & w/o Proto. & 86.97 & 87.25 & 57.42 & 85.99 & 86.95 & 85.92 & 96.80 & 96.84 & 92.74 & 95.92 & 96.99 & 97.41 \\
  & SeVeR      & \textbf{87.85} & \textbf{87.82} & \textbf{87.47} & \textbf{87.61} & \textbf{87.43} & \textbf{87.83} & \textbf{97.78} & \textbf{97.77} & \textbf{97.71} & \textbf{97.73} & \textbf{97.53} & \textbf{97.77} \\ \midrule
Holistic Diagnosis
  & w/o Proto. & 85.19 & 85.56 & 44.62 & 84.39 & 85.33 & 85.70 & 96.61 & 96.71 & 89.30 & 95.69 & 96.76 & 97.53 \\
  & SeVeR      & \textbf{86.05} & \textbf{86.01} & \textbf{85.81} & \textbf{86.07} & \textbf{85.90} & \textbf{85.86} & \textbf{97.59} & \textbf{97.60} & \textbf{97.57} & \textbf{97.60} & \textbf{97.40} & \textbf{97.59} \\ \midrule
Pathology Prediction
  & w/o Proto. & 94.70 & 95.02 & 62.61 & 93.80 & 94.84 & 94.99 & 98.12 & 98.19 & 92.15 & 97.17 & 98.25 & 98.98 \\
  & SeVeR      & \textbf{95.66} & \textbf{95.65} & \textbf{96.00} & \textbf{95.67} & \textbf{95.48} & \textbf{95.53} & \textbf{99.11} & \textbf{99.11} & \textbf{99.16} & \textbf{99.11} & \textbf{98.91} & \textbf{99.10} \\
\bottomrule
\end{tabular}}\vspace{-0.5em}
\caption{Full task performance under different modality settings (S1--S6). \textbf{Top}: MC accuracy (Acc.) and free-text BLEU. \textbf{Bottom}: free-text ROUGE and BERTScore.}
\label{tab:modality_settings_full}\label{tab:modality_acc}\label{tab:modality_open}
\end{table*}

Fig.~\ref{fig:modality_addition} extends the modality-robustness comparison to BLEU, ROUGE, and BERTScore for free-text answers, complementing the accuracy curves in Fig.~\ref{fig:modality_ablation} from the main text. Across all three metrics, \model (solid lines) degrades more gracefully than the w/o Proto.\ baseline (dashed), especially under the most resource-constrained settings. The free-text results thus confirm the same conclusion as the discriminative analysis: prototype-based visual compression yields representations that are robust to missing or substituted MRI sequences.

 \end{document}